\documentclass[fleqn,10pt]{wlscirep}
\usepackage[utf8]{inputenc}
\usepackage[T1]{fontenc}

\usepackage{xcolor}
\usepackage{caption}
\usepackage{amssymb}
\usepackage{booktabs}
\usepackage{bm}
\usepackage{multirow}
\usepackage{diagbox}
\usepackage{tabularray}

\title{Joint Stereo 3D Object Detection and Implicit Surface Reconstruction}

\author[1,*]{Shichao Li}
\author[1,+]{Xijie Huang}
\author[2,+]{Zechun Liu}
\author[1]{Kwang-Ting Cheng}
\affil[1]{Department of Computer Science and Engineering, HKUST, Hong Kong SAR 999077, China}
\affil[2]{Meta Reality Labs, Pittsburgh 15222, USA}

\affil[*]{nicholas.li@connect.ust.hk}

\affil[+]{these authors contributed equally to this work}

\begin{abstract}
	We present a new learning-based framework S-3D-RCNN that can recover accurate object orientation in SO(3) and simultaneously predict implicit rigid shapes from stereo RGB images. For orientation estimation, in contrast to previous studies that map local appearance to observation angles, we propose a progressive approach by extracting meaningful Intermediate Geometrical Representations (IGRs). This approach features a deep model that transforms perceived intensities from one or two views to object part coordinates to achieve direct egocentric object orientation estimation in the camera coordinate system. To further achieve finer description inside 3D bounding boxes, we investigate the implicit shape estimation problem from stereo images. We model visible object surfaces by designing a point-based representation, augmenting IGRs to explicitly address the unseen surface hallucination problem. Extensive experiments validate the effectiveness of the proposed IGRs, and S-3D-RCNN achieves superior 3D scene understanding performance. We also designed new metrics on the KITTI benchmark for our evaluation of implicit shape estimation.
\end{abstract}
\begin{document}
	
	\flushbottom
	\maketitle
	\thispagestyle{empty}
	
	\begin{figure*}[t]
		\begin{center}
			\centering
			\includegraphics[width=\linewidth, trim=0cm 0cm 0cm 2cm]{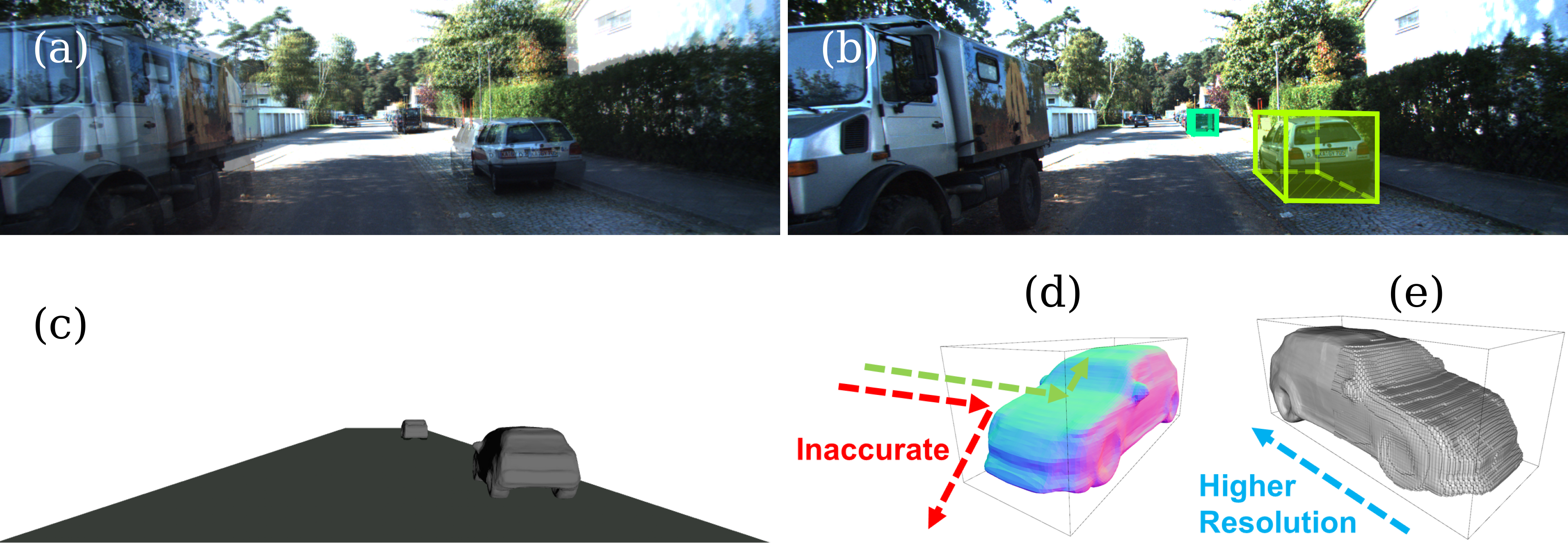}
			\caption{Given a pair of stereo RGB images, S-3D-RCNN can detect 3D objects and predict implicit rigid shapes with one forward pass. (a) Alpha-blended image pair to show the disparities. (b) 3D object proposals shown as 3D bounding boxes. (c) Shape predictions for the detected objects. (d) Estimated surface normal of the nearby object where the red ray indicates incorrect reflection effects with only the 3D bounding box prediction. (e) The predicted implicit shape supports a spatially varying resolution.}
			\label{teaser}
		\end{center}
	\end{figure*}
	
	\section*{Introduction}
	\begin{quote}
		\textit{``The usefulness of a representation depends upon how well suited it is to the purpose for which it is used".}
		\hfill
		--Marr~\cite{marr2010vision}
	\end{quote}
	
	Estimating 3D attributes of outdoor objects is a fundamental vision task enabling many important applications. For example, in vision-based autonomous driving and traffic surveillance systems~\cite{ferryman2000visual}, accurate vehicle orientation estimation (VOE) can imply a driver's intent of travel direction, assist motion prediction and planning, and help identify anomalous behaviors. In outdoor augmented reality systems, rigid shape estimation can enable photo-realistic lighting and physics-based surface effect simulation. In auto-labeling applications~\cite{yang2021auto4d}, off-board deep 3D attribute estimation models serve as an important component in a data closed-loop platform by speeding up the labeling efficiency. This study proposes a new multi-task model to fulfill this task, which takes a pair of calibrated images (Fig.~\ref{teaser}(a)) and can detect objects as bounding boxes (Fig.~\ref{teaser}(b)) as well as estimate their implicit shapes (Fig.~\ref{teaser}(c)). 
	
	Even though the binocular human visual system can recover \emph{multiple} 3D object properties effortlessly from a glance, this task is challenging for computers to accomplish. The difficulty results from a lack of geometry information after image formation, which causes the \emph{semantic gap} between RGB pixels and unknown 3D properties. This problem is even exacerbated by a huge variation in object appearances. Recently, advances in deep learning greatly facilitated the representation learning process from data, where a neural network predicts 3D attributes from images~\cite{mousavian20173d, brazil2019m3d}. In this paradigm, paired images and 3D annotations are specified as inputs and learning targets respectively for supervising a deep model. No intermediate representation is designed in these studies, which need a large number of training pairs to approximate the highly non-linear mapping from the pixel space to 3D geometrical quantities. 
	
	To address this problem, instead of directly regressing them from pixels with a black-box neural network, we propose a progressive mapping from pixels to 3D attributes. This design is inspired by Marr's representational framework of vision~\cite{marr2010vision}. In Marr's framework, intermediate representations, i.e., the $2\frac{1}{2}$-D sketch, are computed from low-level pixels and are later lifted to 3D model representations. However, how to design effective intermediate representations toward accurate and robust outdoor 3D attribute recovery is still task-dependent and under-explored. In this study, our research question is \emph{can a neural network extract explicit geometrical quantities from monocular/stereo RGB images and use them for effective object pose/shape estimation?}

	The conference version of this study~\cite{Li_2021_CVPR} addressed a part of the question, i.e., how to estimate orientation for one class of objects (vehicles) from a single RGB image. The proposed model, Ego-Net, computed part-based screen coordinates from object part heatmaps and further lifted them to 3D object part coordinates for accurate orientation inference. While the study~\cite{Li_2021_CVPR} is effective for VOE, it is limited in several aspects which demands further exploration. Firstly, the approach was demonstrated only for a single-view setting. The difficulty of monocular depth estimation makes it less competent in perception accuracy compared with multi-view systems. Thus, an extension to multi-view sensor configuration and a study of the effectiveness of the proposed IGRs in such cases is highly favorable and complementary. Secondly, the approach was only validated for vehicles and not shown for other common articulated objects such as pedestrians. These objects are smaller and have much fewer training labels in the used dataset. It would be intriguing whether the proposed IGRs show consistent effectiveness or not. Lastly, the approach can only predict object orientation. It does not unlock the full potential of the extracted high-resolution instance features to recover a more detailed rigid shape description, and neither does it discuss how to design effective IGRs to achieve it. A complete and detailed rigid shape description beyond 3D bounding boxes is desirable for various machine vision applications. For example, an autonomous perception system can give a more accurate collision prediction \emph{within} an object's 3D bounding box. As an illustration, the green ray reflecting on the predicted object surface in Fig.~\ref{teaser}(d) cannot be obtained by using 3D bounding boxes to represent objects due to the lack of fine-grained surface normal. In addition, our approach describes rigid objects with implicit representations, which can be rendered with varying resolutions (Fig.~\ref{teaser}(e)).
	
	To fully address our research question, this study presents an extended model S-3D-RCNN for joint object detection, orientation estimation, and implicit shape estimation from a pair of RGB images. Firstly, we demonstrate the effectiveness of the proposed IGRs in a stereo perception setting, where part-based screen coordinates aggregated from two views further improve the VOE accuracy. Secondly, we validate the robustness of the proposed IGRs for other outdoor objects such as pedestrians. These tiny objects are underrepresented in the training dataset, yet the proposed approach still achieves accurate orientation estimation. Lastly, we propose several new representations in the framework to further extend EgoNet for implicit shape estimation. We formulate the problem of implicit shape estimation as an unseen surface hallucination problem and propose to address it with a point-based visible surface representation. For quantitative evaluation, we further propose two new metrics to extend KITTI's object detection evaluation~\cite{geiger2012we} to consider the retrieval of the object surface. In summary, this study extends the previous conference version in various aspects with the following added contributions.
	
	\begin{itemize}
		\item It explores the proposed IGRs in a stereo-perception setting and validates the effectiveness of the proposed approach in recovering orientation in SO(3) for two-view inputs.
		\item It shows the proposed approach is not limited to rigid objects and has strong orientation estimation performance for other small objects that may have fewer training labels.
		\item It extends the representational framework in Ego-Net with several new IGRs to achieve implicit shape estimation from stereo region-of-interests. To the best of our knowledge, Ego-Net++ is the first stereo image-based approach for 3D object detection and implicit rigid shape estimation.
		\item To quantitatively evaluate the proposed implicit shape estimation task, two new metrics are designed to extend the previous average precision metrics to consider the object surface description.
	\end{itemize}
	
	We introduce and compare with relevant studies in the next section and revisit Ego-Net in section~\ref{Ego-Net}. We then detail extended studies in designing Ego-Net++ in senction~\ref{sec:egonet++} followed by experimental evaluations in section~\ref{sec:experiments}.

	\section{Related Work}
	This study features outdoor environment, orientation estimation, and implicit shape reconstruction. It draws connections with prior studies in the following domains yet has unique contributions.
	
	\noindent \textbf{Image-based 3D scene understanding} requires recovering 3D object properties from RGB images which usually consist of \emph{multiple} sub-tasks~\cite{hoiem2008closing, geiger2011joint, kim20133d, zhang2014panocontext, tulsiani2018factoring, chen2019holistic++, hampali2021monte, dahnert2021panoptic}. Two popular paradigms were proposed. The \emph{generative}, a.k.a \emph{analysis-by-synthesis} approaches~\cite{yuille2006vision, loper2014opendr, niemeyer2020differentiable, zakharov2021singleshot} build generative models of image observation and unknown 3D attributes. During inference, they search in the 3D state space to find an optimum that best explains the image evidence. However, good initialization and iterative optimization are required to search in a high-dimensional state space. In contrast, the \emph{discriminative} approaches~\cite{eslami2016attend, chabot2017deep, kundu20183d, engelmann2021points} directly learn a mapping from image observation to 3D representations. Our approach can be categorized into the latter yet is unique. Unlike previous studies that are only applicable for indoor environments with small depth variation~\cite{schwing2012efficient, zhang2014panocontext,nie2020total3dunderstanding, runz2020frodo, zhang2021deeppanocontext, liu2021voxel} or only consider the monocular camera setting~\cite{huang2018holistic, kundu20183d, gkioxari2019mesh, mustafa2021multi}, our framework can exploit two-view geometry to accurately locate objects as well as enables resolution-agnostic implicit shape estimation in challenging outdoor environments. Compared with recent multi-view reconstruction studies~\cite{10155236, 9711318}, our study does not take 3D mesh inputs as~\cite{10155236} or use synthetic image inputs as~\cite{9711318}.
	
	\noindent \textbf{Learning-based 3D object detection} learns a function that maps sensor input to objects represented as 3D bounding boxes~\cite{3dop}. Depending on the sensor configuration, previous studies can be categorized into RGB-based methods~\cite{brazil2019m3d, disentangling, zhou2020iafa, Li_2021_CVPR, reading2021categorical, lian2022exploring, chen2022pseudo} and LiDAR-based approaches~\cite{second, zhou2018voxelnet, shi2019pointrcnn, shi2021pv}. Our approach is RGB-based which does not require expensive range sensors. While previous RGB-based methods can describe objects up to a 3D bounding box representation, the quality of shape predictions within the bounding boxes was not evaluated with existing metrics. Our extended study fills this gap by proposing new metrics and designing an extended model that complements Ego-Net with implicit shape reconstruction capability from stereo inputs.
	
	\noindent \textbf{Learning-based orientation estimation for 3D object detection} seeks a function that maps pixels to instance orientation in the camera coordinate system via learning from data. Early studies~\cite{juranek2015real, xiang2015data} utilized hand-crafted features~\cite{dollar2014fast} and boosted trees for discrete pose classification (DPC). More recent studies replace the feature extraction stage with deep models. ANN~\cite{yang2014object} and DAVE~\cite{zhou2016dave, zhou2017fast} classify instance feature maps extracted by CNN into discrete bins. To deal with images containing multiple instances, Fast-RCNN-like architectures were employed in~\cite{3dop, braun2016pose, 3dop-pami, huang2019perspectivenet, ke2020gsnet} where region-of-interest (ROI) features were used to represent instance appearance and a classification head gives pose prediction. Deep3DBox~\cite{mousavian20173d} proposed \emph{MultiBin} loss for joint pose classification and residual regression. Wasserstein loss was promoted in~\cite{liu2019conservative} for DPC. Our Ego-Net~\cite{Li_2021_CVPR} is also a learning-based approach but possesses key differences. Our approach promotes learning explicit part-based IGRs while previous works do not. With IGRs, Ego-Net is robust to occlusion and can directly estimate global (egocentric) pose in the camera coordinate system while previous works can only estimate relative (allocentric) pose. Compared to Ego-Net, orientation estimation with Ego-Net++ in this extended study is no longer limited to monocular inputs and rigid objects. In addition, Ego-Net++ can further achieve object surface retrieval for rigid objects while Ego-Net cannot. 
	
	\noindent \textbf{Instance-level modeling in 3D object detection} builds a feature representation for a single object to estimate its 3D attributes~\cite{kundu20183d, liu2019deep, peng2020ida, liu2020reinforced, ke2020gsnet}. FQ-Net~\cite{liu2019deep} draws a re-projected 3D cuboid on an instance patch to predict its 3D Intersection over Union (IoU) with the ground truth. RAR-Net~\cite{liu2020reinforced} formulates a reinforcement learning framework for instance location prediction. 3D-RCNN~\cite{kundu20183d} and GSNet~\cite{ke2020gsnet} learn a mapping from instance features to the PCA-based shape codes. Ego-Net++ in this study is a new instance-level model in that it can predict the implicit shape and can utilize stereo imagery while previous studies cannot.
	
	\noindent \textbf{Neural implicit shape representation} was proposed to encode object shapes as latent vectors via a neural network~\cite{park2019deepsdf, mescheder2019occupancy, chabra2020deep, erler2020points2surf, takikawa2021neural}, which shows an advantage over classical shape representations. However, many prior works focus on using perfect synthetic point clouds as inputs and few have explored its inference in 3D object detection (3DOD) scenarios. Instead, we address inferring such representations under a realistic object detection scenario with stereo sensors by extending the IGRs in Ego-Net to accomplish this task. 
	
	\section{Method}
	\subsection{Overall framework}
	Our framework S-3D-RCNN detects objects and estimates their 3D attributes from a pair of stereo images with designed intermediate representations. S-3D-RCNN consists of a proposal model $\mathcal{D}$ and an instance-level model $\mathcal{E}$ (Ego-Net++) for instance-level 3D attribute recovery as shown in Fig.~\ref{fig:egonet++}. $\mathcal{E}$ is agnostic to the design choice of $\mathcal{D}$ and can be used as a plug-and-play module. Given an image pair $(\mathcal{L}, \mathcal{R})$ captured by stereo cameras with left camera intrinsics $\text{K}_{3 \times 3}$, $\mathcal{D}$ predicts $N$ cuboid proposals $\{ b_i \}_{i=1}^{N}$ as
	$
	\mathcal{D}(\mathcal{L}, \mathcal{R}; \theta_{\mathcal{D}}) = \{ b_q \}_{q=1.}^{N}
	$
	Conditioned on each proposal $b_i$, $\mathcal{E}$ constructs instance-level representations and predicts its orientation as    
	$
	\mathcal{E}(\mathcal{L}, \mathcal{R}; \theta_{\mathcal{E}}|b_i) = \bm{\theta}_{i.}
	$
	In addition, for rigid object proposals (i.e., vehicles in this study), $\mathcal{E}$ can further predict its implicit shape representation. In implementation, $\mathcal{D}$ is designed as a voxel-based 3D object detector as shown in Fig.~\ref{global_network}. The following sub-sections first revisit the motivation and representation design in Ego-Net, and then highlight which new representations are introduced in Ego-Net++ for a stronger 3D scene understanding performance. 
	\begin{figure*}
		\begin{center}
			\centering
			\includegraphics[width=\linewidth, trim=0cm 0cm 0cm 0cm]{./figs/ego-net++.jpg}
			\captionof{figure}{Diagram of $\mathcal{E}$ (Ego-Net++). $\mathcal{E}$ performs orientation and rigid shape estimation with intermediate geometric representations. A local cost volume is constructed from instance features to estimate disparities. The visible surface coordinates are computed from the predicted disparities and an estimated mask, and then normalized to a canonical coordinate system. An encoder-decoder component $Ha$ infers the missing surface of the object. The complete surface coordinates are passed to an encoder to extract an implicit shape vector, which can be used by a decoder for resolution-agnostic mesh extraction. For orientation estimation, a zoomed-in view is shown in Fig.~\ref{fig:egonet++_orientaion}. FCN stands for a fully convolutional network module. The 2D part coordinates are lifted to 3D coordinates by $Li$ in Eq.~\ref{eq:comp_graph_ego_net++}.}
			\label{fig:egonet++}
		\end{center}
	\end{figure*} 
	
	\begin{figure*}
		\begin{center}
			\centering
			\includegraphics[width=\linewidth, trim=0cm 0cm 0cm 0cm]{./figs/proposal_model.jpg}
			\captionof{figure}{The proposal model $\mathcal{D}$ of S-3D-RCNN. In this implementation, a volumetric 3D scene representation is built from semantic features and cost-volume-based geometric features similar to~\cite{chen2020dsgn}. An anchor-based object detector processes the Bird's Eye View feature maps to generate 3D object proposals. Note that Ego-Net++ $\mathcal{E}$ is agnostic to the design choice of $\mathcal{D}$ and can be used with other 3D object detectors.}
			\label{global_network}
		\end{center}
	\end{figure*}
	\subsection{Ego-Net: monocular egocentric vehicle pose estimation with IGRs}
	\label{Ego-Net}
	\subsubsection{Orientation estimation with a progressive mapping}
\begin{figure}[h]
		\begin{center}
			\includegraphics[width=0.5\linewidth, trim=4cm 4cm 4cm 5cm]{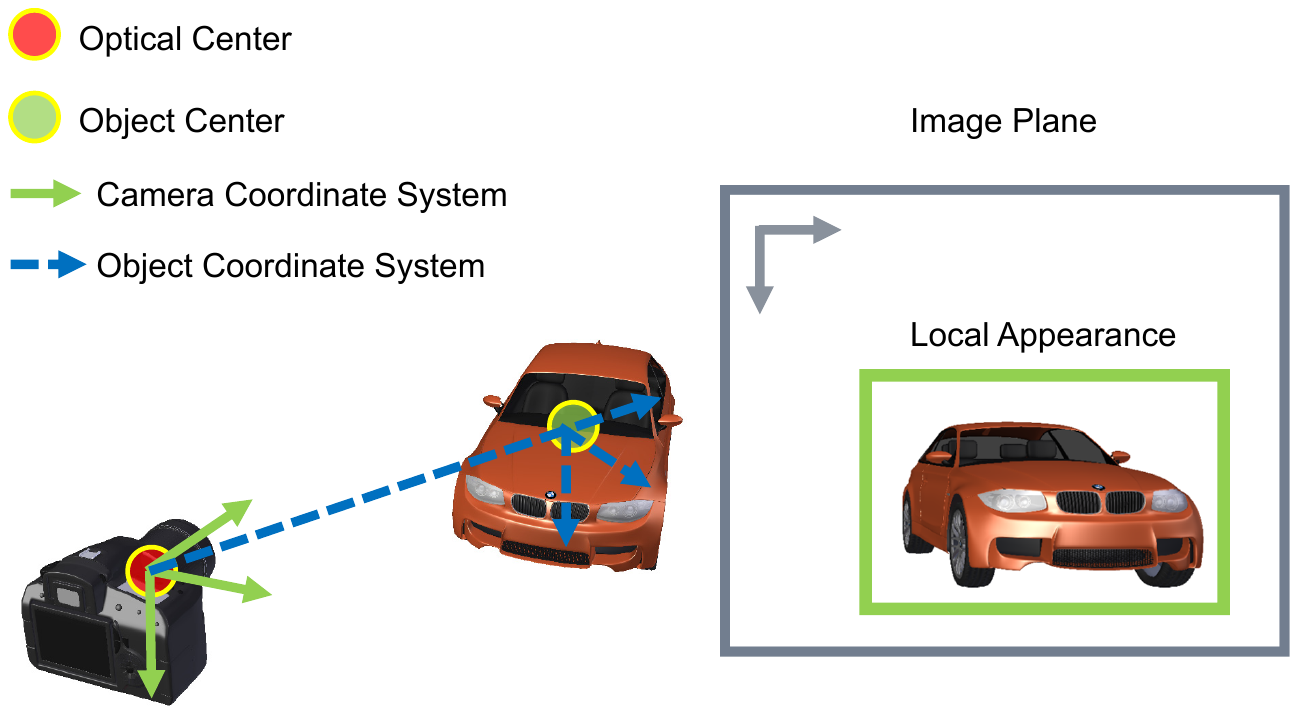}
		\end{center}
		\caption{Local appearance cannot uniquely determine egocentric pose. Existing solutions first estimate an allocentric pose in the object coordinate system (blue) and convert it to an egocentric pose in the camera coordinate system (green) based on the object location.}
		\label{fig:3.1}
	\end{figure}

	Previous studies~\cite{mousavian20173d, kundu20183d, brazil2019m3d, Ding_2020_CVPR} regress vehicle orientation with the computational graph in Eq.~\ref{eq:comp_graph_previous}. A CNN-based model $\mathcal{N}$ is used to map local instance appearance $\mathbf{x}_i$ to allocentric pose, i.e., 3D orientation in the object coordinate system (OCS), which is then converted to the egocentric pose, i.e., orientation in the camera coordinate system (CCS). The difference between these two coordinate systems is shown in Fig.~\ref{fig:3.1}. This two-step design is a workaround since an object with the same egocentric pose $\bm{\theta}_i$ can produce different local appearance depending on its location~\cite{kundu20183d} and learning the mapping from $\mathbf{x}_i$ to $\bm{\theta}_i$ is ill-posed. In this two-step design, OCS was estimated by another module. The error of this module can propagate to the final estimation of egocentric poses, and optimizing $\mathcal{N}$ does not optimize the final target directly. Another problem of this design is that the mapping from pixels $\mathbf{x}_i$ to pose vectors $\bm{\alpha}_i$ is highly non-linear and difficult to approximate~\cite{tompson2014joint}.
	\begin{equation}
	\label{eq:comp_graph_previous}
	\begin{array}{llll}
	\mathbf{x}_i  & \xrightarrow{\mathcal{N}} & ~~ \bm{\alpha}_i & \xrightarrow{convert} ~~ \bm{\theta}_i \\
	& & \text{OCS} & \nearrow{}\\
	\end{array}
	\end{equation}
	
	\begin{figure*}
		\begin{center}
			\includegraphics[width=0.8\linewidth, trim=2cm 0cm 13cm 2cm]{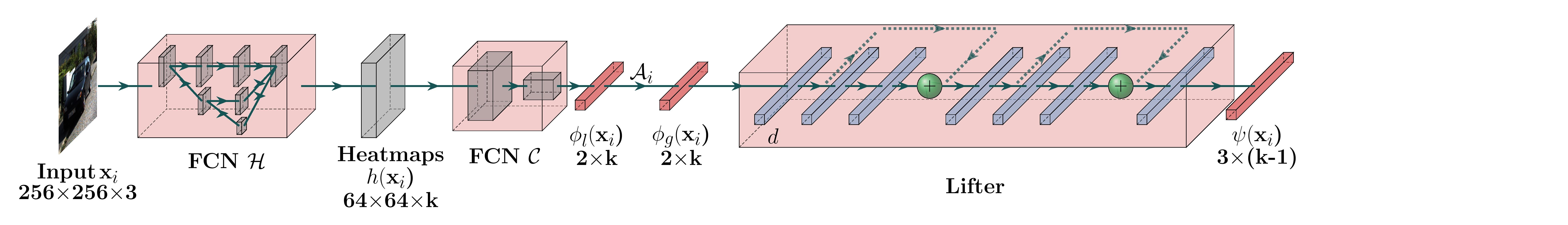}
		\end{center}
		\caption{Model architecture of Ego-Net. A fully convolutional model $\mathcal{H}$ regresses part heatmaps from a 2D patch of a proposal. The heatmaps are mapped to local coordinates with several strided convolution layers. The local coordinates are transformed to screen coordinates $\phi_g(\mathbf{x}_i)$ and mapped to a point-based 3D representation $\psi(\mathbf{x}_i)$ of a cuboid, whose orientation directly represents egocentric pose in the camera coordinate system. k=33 when q=2 as in Sec.~\ref{sec:igrs}.}
		\label{fig:egonet}
	\end{figure*}
	
	Ego-Net instead learns a mapping from images to egocentric poses to optimize the target directly. However, instead of relying on a black box model to fit such a non-linear mapping, it promotes a progressive mapping, where coordinate-based IGRs are extracted from pixels and eventually lifted to the 3D target. Specifically, Ego-Net is a composite function with learnable modules $\{\mathcal{H},\mathcal{C}, Li\}$. Given the cropped 2D image patch of one proposal $\mathbf{x}_i$, Ego-Net predicts its egocentric pose as $\mathcal{E}(\mathbf{x}_i) = Li(\mathcal{A}_i(\mathcal{C}(\mathcal{H}(\mathbf{x}_i)))) =  \bm{\theta}_i$. Fig.~\ref{fig:egonet} depicts Ego-Net, whose computational graph is shown in Eq.~\ref{eq:comp_graph_ego_net}. $\mathcal{H}$ extracts heatmaps $h(\mathbf{x}_i)$ for 2D object parts that are mapped by $\mathcal{C}$ to coordinates $\phi_l(\mathbf{x}_i)$ representing their local location on the patch. $\phi_l(\mathbf{x}_i)$ is converted to the global image plane coordinates $\phi_g(\mathbf{x}_i)$ with an affine transformation $\mathcal{A}_i$ parametrized with scaling and 2D translation. $\phi_g(\mathbf{x}_i)$ is further lifted to a 3D representation $\psi(\mathbf{x}_i)$ by $Li$. The final pose prediction derives from $\psi(\mathbf{x}_i)$.        
	\begin{equation}
	\label{eq:comp_graph_ego_net}
	\arraycolsep=1pt
	\begin{array}{lllllllllll}
	\mathbf{x}_i  & \xrightarrow{\mathcal{H}} & h(\mathbf{x}_i) & \xrightarrow{\mathcal{C}} & \phi_l(\mathbf{x}_i) & \xrightarrow{\mathcal{A}_i} & \phi_g(\mathbf{x}_i) & \xrightarrow{Li} & \psi(\mathbf{x}_i)  & \rightarrow & \bm{\theta}_i
	\end{array}
	\end{equation} 
	
	\newcommand{\visi}{v(\mathbf{x}_i)}%
	\newcommand{\ocsi}{o(\mathbf{x}_i)} 
	\newcommand{\compi}{c(\mathbf{x}_i)}%
	\newcommand{\shapei}{s(\mathbf{x}_i)}%
	\newcommand{\maski}{\mathcal{M}(\mathbf{x}_i)}%
	\newcommand{\masks}{\mathcal{M}^{sp}(\mathbf{x}_i)}%
	
	\subsubsection{Design of Labor-free Intermediate Representations}
	\label{sec:igrs}
	The IGRs in Eq.~\ref{eq:comp_graph_ego_net} are designed based on the following considerations:
	
	\noindent \emph{Availability}: It is favorable if the IGRs can be easily derived from existing ground truth annotations with none or minimum extra manual effort. Thus we define object parts from existing 3D bounding box annotations.
	
	\noindent \emph{Discriminative}: The IGRs should be indicative for orientation estimation, so that they can serve as a good bridge between visual appearance input and the geometrical target.
	
	\noindent \emph{Transparency}: The IGRs should be easy to understand, which makes them debugging-friendly and trustworthy for applications such as autonomous driving. Thus IGRs are defined with explicit meaning in Ego-Net.
	
\begin{figure}[t]
		\begin{center}
			\includegraphics[width=0.4\linewidth, trim=8cm 4cm 8cm 3cm]{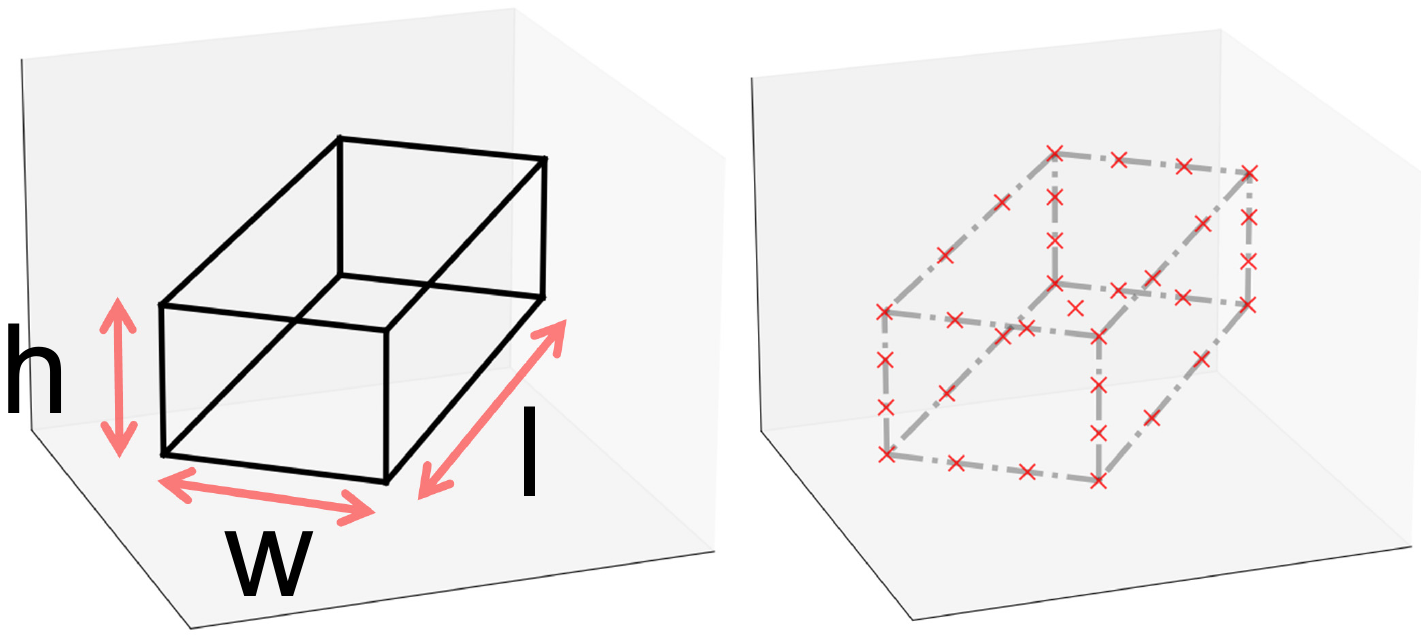}
		\end{center}
		\caption{An instance $\mathbf{x}_i$ with primitive 3D bounding box annotation (left) and its 3D point cloud representation $\psi(\mathbf{x}_i)$ (right) in the camera coordinate system as the collection of red crosses.}
		\label{fig:3drep}
	\end{figure}

	With the above considerations, we define the 3D representation  $\psi(\mathbf{x}_i)$ as a sparse 3D point cloud (PC) representing an interpolated cuboid. Autonomous driving datasets such as KITTI~\cite{geiger2012we} usually label instance 3D bounding boxes from captured point clouds where an instance $\mathbf{x}_i$ is associated with its centroid location in the camera coordinate system $\mathbf{t}_i = [t_x, t_y, t_z]$, size $[h_i, w_i, l_i]$, and its egocentric pose $\bm{\theta}_i$. For consistency, many prior studies only use the yaw angle denoted as $\theta_i$. As shown in Fig.~\ref{fig:3drep}, denote the 12 lines comprising a 3D bounding box as $\{\mathbf{l}_j\}_{j=1}^{12}$, where each line is represented by two endpoints (\textbf{s}tart and \textbf{e}nd) as $\mathbf{l}_j = [\mathbf{p}_j^{s}; \mathbf{p}_j^{e}]$. $\mathbf{p}_j^v$  ($v$ is $s$ or $e$) is a 3-vector $(X_j^{v}, Y_j^{v}, Z_j^{v})$ representing the point's location in the camera coordinate system. As a complexity-controlling parameter, $q$ more points are sampled from each line with a pre-defined interpolation matrix $B_{q \times 2}$ as
	\begin{equation}
	\begin{bmatrix}
	\mathbf{p}_j^1 \\
	\mathbf{p}_j^2 \\
	\dots \\
	\mathbf{p}_j^q \\
	\end{bmatrix}
	=
	B_{q \times 2}
	\begin{bmatrix}
	\mathbf{p}_j^{s} \\
	\mathbf{p}_j^{e} \\
	\end{bmatrix}
	=
	\begin{bmatrix}
	\beta_1 & 1 - \beta_1 \\
	\beta_2 & 1 - \beta_2 \\
	\dots & \dots \\
	\beta_q & 1 - \beta_q \\
	\end{bmatrix}
	\begin{bmatrix}
	\mathbf{p}_j^{s} \\
	\mathbf{p}_j^{e} \\
	\end{bmatrix}.
	\end{equation} 
	The 8 endpoints, the instance's centroid, and the interpolated points for each of the 12 lines form a set of $9 + 12q$ points. The concatenation of these points forms a $9 + 12q $ by $3$ matrix $\tau(\mathbf{x}_i)$. Since we do not need the 3D target $\psi(\mathbf{x}_i)$ to encode location, we deduct the instance translation $\mathbf{t}_i$ from $\tau(\mathbf{x}_i)$ and represent $\psi(\mathbf{x}_i)$ as a set of $8 + 12q$ points representing the shape relative to the centroid  
	$
	\psi(\mathbf{x}_i) = \{(X_{j}^{v} - t_x, Y_{j}^{v} - t_y, Z_{j}^{v} - t_z)\}
	$
	where $v\in\{s,1,\dots,q,e\}$ and $j\in\{1,2,\dots,12\}$.
	Larger $q$ provides more cues for inferring pose yet increases complexity. In practice, we choose $q=2$ and the right figure of Fig.~\ref{fig:3drep} shows an example with 
	$B_{2 \times 2}= \begin{bmatrix}
	\frac{3}{4} & \frac{1}{4} \\
	\frac{1}{4} & \frac{3}{4} \\
	\end{bmatrix}
	$
	and 2 points are interpolated for each line.
	
	Serving as the 2D representation to be located by $\mathcal{H}$ and $\mathcal{C}$, $\phi_g(\mathbf{x}_i)$ is defined to be the projected \emph{screen coordinates} of $\tau(\mathbf{x}_i)$ given camera intrinsics $\text{K}_{3 \times 3}$ as
	\begin{equation}
	\phi_g(\mathbf{x}_i) = \text{K}_{3 \times 3} \tau(\mathbf{x}_i).
	\end{equation}
	$\phi_g(\mathbf{x}_i)$ implicitly encodes instance location on the image plane so that it is less ill-posed to estimate egocentric pose from it directly. In summary, these IGRs can be computed with zero extra manual annotation, are easy to understand, and contain rich information for estimating the instance orientation.
	
	\subsection{Ego-Net++: towards multi-view perception and implicit shape inference}
	To study the effectiveness of \emph{screen coordinates} in encoding two-view information, and achieve a finer description for perceived objects, several new IGRs are designed in Ego-Net++.
	\label{sec:egonet++}
	\subsubsection{Orientation estimation with paired part coordinates}
	Under the stereo perception setting, another viewpoint provides more information to infer the unknown object orientation. It is thus necessary to extend the IGRs to aggregate information from both views.%
	\begin{figure}[h!]
		\begin{center}
			\includegraphics[width=0.6\linewidth]{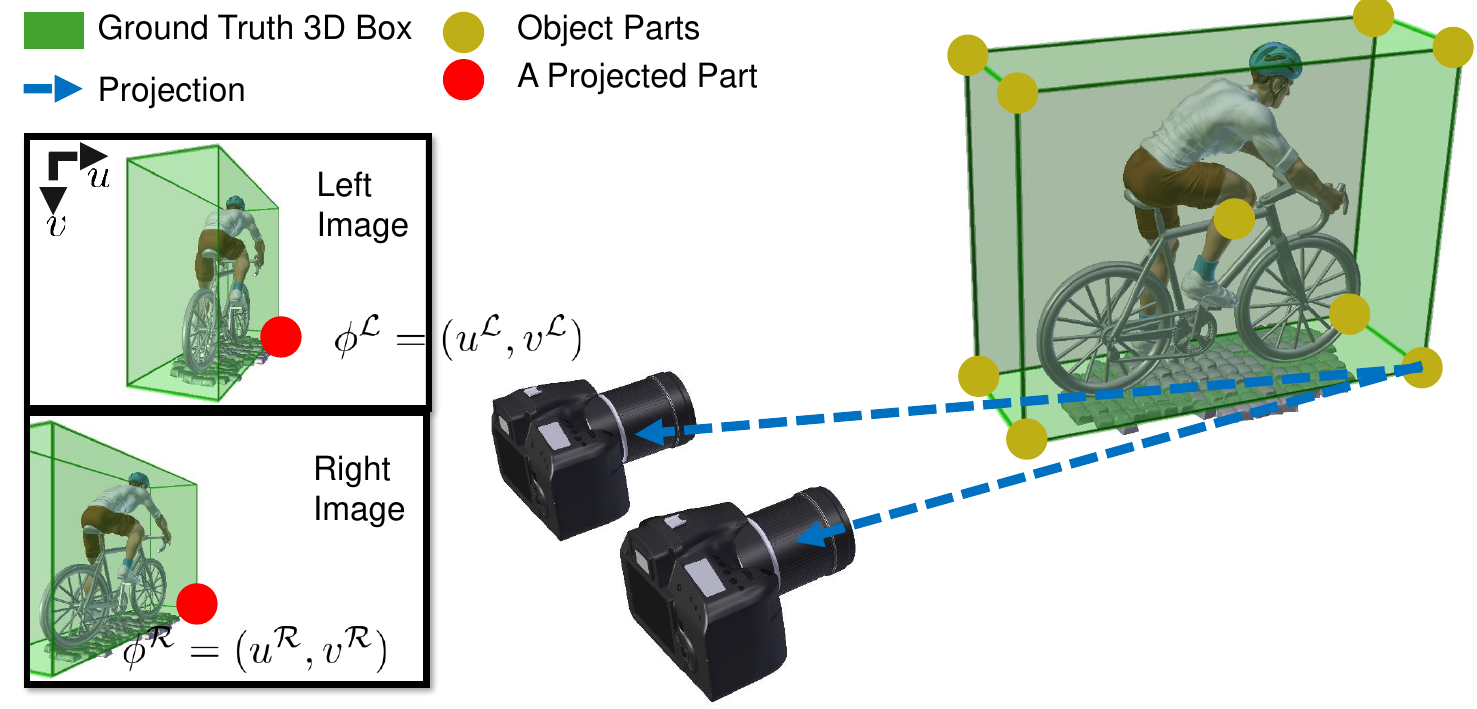}
		\end{center}
		\caption{Diagram of the \emph{paired part coordinates} representation (red) for an object part of a non-rigid object.}
		\label{ppc}
	\end{figure}
	\begin{figure}[h]
		\begin{center}
			\includegraphics[width=0.6\linewidth]{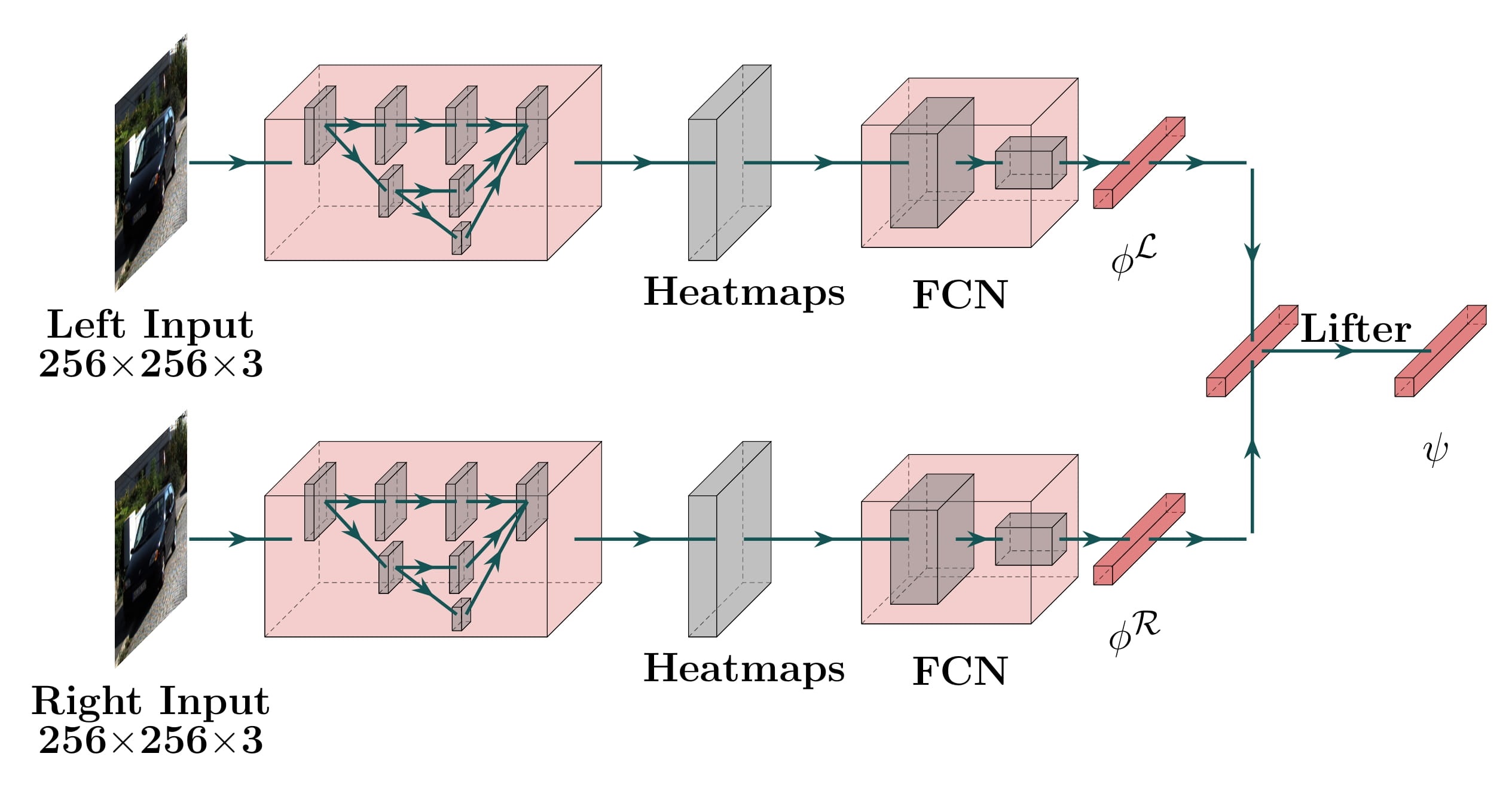}
		\end{center}
		\caption{Zoomed-in view of the orientation estimation part in $\mathcal{E}$ in Fig.~\ref{fig:egonet++}.}
		\label{fig:egonet++_orientaion}
	\end{figure}
	In Ego-Net++, \emph{paired part coordinates} (PPCs) are defined with a simple yet effective concatenation operation to aggregate two-view information as a k by 4 representation 
	\begin{equation}
	\phi_g(\mathbf{x}_i^{\mathcal{L}}, \mathbf{x}_i^{\mathcal{R}}) = \phi_g(\mathbf{x}_i^{\mathcal{L}}) \oplus \phi_g(\mathbf{x}_i^{\mathcal{R}}),	
	\end{equation}
	PPC enhances the IGR in Ego-Net with disparity, i.e., differences of part coordinates in two views. Such disparity provides geometrical cues for object part depth and have larger mutual information with the target orientation. PPC is illustrated in Fig.~\ref{ppc}. for one object part, and it extends the computational graph in Eq.~\ref{eq:comp_graph_ego_net} as
	
	\begin{equation}
	\label{eq:comp_graph_ego_net++}
	\arraycolsep=1pt
	\resizebox{0.48\textwidth}{!}{
		$
		\begin{array}{lllllllllllll}
		\mathbf{x}_i^{\mathcal{L}}  & \xrightarrow{\mathcal{H}} & h(\mathbf{x}_i^{\mathcal{L}}) & \xrightarrow{\mathcal{C}} & \phi_l(\mathbf{x}_i^{\mathcal{L}}) & \xrightarrow{\mathcal{A}_i^{\mathcal{L}}} & \phi_g(\mathbf{x}_i^{\mathcal{L}}) & \xrightarrow{\oplus} & \phi_g(\mathbf{x}_i^{\mathcal{L}}, \mathbf{x}_i^{\mathcal{R}}) & \xrightarrow{Li} & \psi(\mathbf{x}_i)  & \rightarrow & \bm{\theta}_i \\
		\mathbf{x}_i^{\mathcal{R}}  & \xrightarrow{\mathcal{H}} & h(\mathbf{x}_i^{\mathcal{R}}) & \xrightarrow{\mathcal{C}} & \phi_l(\mathbf{x}_i^{\mathcal{R}}) & \xrightarrow{\mathcal{A}_i^{\mathcal{R}}} & \phi_g(\mathbf{x}_i^{\mathcal{R}}) & \nearrow &   &  &  & &	
		\end{array}
		$
	}
	\end{equation}
	
	The learnable modules are implemented with convolutional and fully-connected layers, and a diagram of orientation estimation from stereo images is shown in Fig.~\ref{fig:egonet++_orientaion}. During training, ground truth IGRs are available to penalize the predicted heatmaps, predicted 2D coordinates, and 3D coordinates. Such supervision is implemented with $L_2$ loss for heatmaps and $L_1$ loss for coordinates. In inference, the predicted egocentric pose is computed from the predicted 3D coordinates $\psi(\mathbf{x}_i)$. Denote the 3D coordinates at a canonical pose as $\psi^{0}(\mathbf{x}_i)$, the predicted pose is obtained by estimating a rotation $R(\bm{\theta}_i)$ from $\psi^{0}(\mathbf{x}_i)$ to $\psi(\mathbf{x}_i)$. 
	\begin{equation}
	\label{transform}
	\bm{\theta}_i =\arg\min_{\bm{\theta}_i}||R(\bm{\theta}_i)\psi^{0}(\mathbf{x}_i) - \psi(\mathbf{x}_i)||,
	\end{equation}
	In implementation, this least-square problem is solved with singular value decomposition (SVD). This process is efficient due to a small number of parts.
	
	\subsubsection{Implicit shape estimation via surface hallucination}
	Previous sub-sections enable recovering object orientation from one or two views. However, such perception capability is limited to describing objects as 3D bounding boxes, failing at a more detailed representation within the box. While some previous studies~\cite{zeeshan2014cars, engelmann2016joint, ke2020gsnet, chen2021shape} explore shape estimation for outdoor rigid objects, they cannot exploit the stereo inputs or require PCA-based templates~\cite{zeeshan2014cars, engelmann2016joint, ke2020gsnet, chen2021shape} which are limited to a fixed mesh topology~\cite{ke2020gsnet}. In contrast, Ego-Net++ can take advantage of stereo information and conduct implicit shape estimation which can produce flexible resolution-agnostic meshes. To the best of our knowledge, how to design effective intermediate representations for recovering implicit rigid shapes for outdoor objects with stereo cameras is under-explored.
	
	We design the IGRs for implicit shape estimation based on the following fact. The implicit shape representation $s(\mathbf{x}_i)$ for each rigid object describes its complete surface geometry. However, the observation from stereo cameras only encodes a portion of the object's surface. This indicates that the implicit shape reconstruction problem can be modeled as an \emph{unseen surface hallucination} problem, i.e., one needs to infer the unseen surface based on partial visual evidence. Specifically, Ego-Net++ addresses this problem by extending the IGRs in Ego-Net with several new representations and learning a progressive mapping from stereo appearances to $s(\mathbf{x}_i)$. This mapping is represented as a composite function
	$
	E \circ Ha \circ O \circ V,
	\label{composite}
	$ that has learnable parameters $\{Ha, V, E\}$ and the following computational graph
	\begin{equation}
	\label{eq:comp_graph_implicit_shape}
	\arraycolsep=1pt
	\begin{array}{lllllllll}
	\mathbf{x}_i^{\mathcal{L}}, \mathbf{x}_i^{\mathcal{R}}  & \xrightarrow{V} & v(\mathbf{x}_i) & \xrightarrow{O} & o(\mathbf{x}_i) & \xrightarrow{Ha} & \compi & \xrightarrow{E} & s(\mathbf{x}_i).
	\end{array}
	\end{equation} 
	
	Here $v(\mathbf{x}_i)$ represents the visible object surface. After a normalization operator $O$, such representation is converted to the OCS. To recover the missing surface, a point-based encoder-decoder $Ha$ hallucinates a complete surface based on learned prior knowledge. $E$ encodes the complete shape into an implicit shape vector.
	
	\subsubsection{The visible-surface representation} 
	We propose a point-based representation for the visible object surface. Given a pair of stereo RoIs, $V$ estimates the foreground mask and depth, samples a set of pixels from the foreground, and re-projects them to the CCS. Denote $\maski$ as the predicted set of foreground pixels, we sample $e$ elements from it as $\masks$. These elements are re-projected to the CCS to form a set of $e$ 3D points as
	\begin{equation}
	\{\text{K}_{3 \times 3}^{-1}
	\begin{bmatrix}
	m_z*m^x \\
	m_z*m^y \\
	m_z \\
	\end{bmatrix}\vert
	m \in \masks  \subset \maski 
	\},
	\end{equation}
	where $(m^x, m^y)$ denotes the screen coordinates of pixel $m$. Concatenating these elements gives a 3 by $e$ matrix $\visi$ encoding the visible instance PC in the CCS.
	
	In implementation, $\maski$ is obtained by applying fully convolutional layers using 2D features for foreground classification. To obtain the depth prediction for the foreground region, a local cost volume is constructed to estimate disparity for the local patch. The disparities are then converted to depth as $m^z = fB/m^{disp}$ where $m^{disp}$, $B$, and $f$ are the estimated disparity, stereo baseline length, and focal length respectively. 
	
	\subsubsection{Hallucination with normalized coordinates} We found the learning of unseen surface hallucination more difficult when the input point coordinates represent a large variation of object pose and size. Thus we propose to disentangle the estimation of rigid shape from object pose. Specifically, we use operator $O$ to normalize the visible object surface to a canonical OCS with a similarity transformation. Denote a detected object $b_i$ as a 7-tuple $b_i = (x_i, y_i, z_i, h_i, w_i, l_i, \theta_i)$, where $(x_i, y_i, z_i)$, $(h_i, w_i, l_i)$ and $\theta_i$ denote its translation, size (height, width, length) and orientation in the CCS respectively. The normalized coordinates are computed conditioned on $b_i$ as 
	\begin{equation}
	\ocsi = 
	\begin{gathered}
	\begin{bmatrix}
	cos\theta_i*l_i & 0 & sin\theta_i*l_i & t_x \\
	0 & l_i& 0& t_y \\
	-sin\theta_i*l_i & 0& cos\theta_i*l_i& t_z \\
	0 & 0& 0& 1
	\end{bmatrix}
	^{-1}
	\begin{bmatrix}
	\vline \\
	\visi \\
	\vline \\
	\textbf{1}
	\end{bmatrix}.
	\end{gathered}
	\end{equation}
	$Ha$ is implemented as a point-based encoder-decoder module, which extracts point-level features~\cite{qi2017pointnet} from $\ocsi$ and infers $N_c$ by 3 coordinates $\compi$ to represent the complete surface.
	
	Finally, the shape encoder $E$ maps the estimated complete surface $\compi$ into a latent vector $\shapei$ that encodes the object's implicit shape. To extract a mesh representation from the predicted implicit shape code, $\mathcal{E}$ uses the occupancy decoder~\cite{mescheder2019occupancy} where a set of grid locations is specified and predicts the occupancy field on such grid. Note this grid is not necessarily evenly distributed thus one can easily use the shape code in a resolution-agnostic manner. Given the occupancy field, we then use the Marching Cube algorithm~\cite{lorensen1987marching} to extract an isosurface as a mesh representation.
	
	To optimize the learnable parameters during training, the supervision consists of the cross-entropy segmentation loss, the smooth $L_1$ disparity estimation loss, and the hallucination loss implemented as Chamfer distance. We train the shape decoder on ShapeNet~\cite{chang2015shapenet} by randomly sampling grid locations within object 3D bounding boxes with paired ground truth occupancy. In inference, we apply zero padding if $Card(\maski)<e$.  
	
	\subsubsection{Penalizing in-box descriptions for 3DOD}
	Per the new IGRs introduced in Ego-Net++, S-3D-RCNN can estimate 3D bounding boxes accurately from stereo cameras as well as describe a finer rigid shape \emph{within} the bounding boxes. However, existing 3DOD studies on KITTI~\cite{geiger2012we} cannot measure the goodness of a more detailed shape beyond a 3D bounding box representation. To fill this gap and validate the effectiveness of the new IGRs, we contribute new metrics to KITTI for the intended evaluation.
	
	As a reference metric, the official \emph{Average Orientation Similarity} ($AOS$) metric in KITTI is defined as  
	\begin{equation}
	AOS = \frac{1}{11}\sum_{r\in\{0, 0.1, \dots, 1\}}\text{max}_{\tilde{r}:\tilde{r}\geq r}s(\tilde{r}),
	\label{eq:def_aos}
	\end{equation}
	where $r$ is the detection recall and $s(r) \in [0, 1]$ is the orientation similarity (OS) at recall level $r$. OS is defined as 
	\begin{equation}
	s(r) = \frac{1}{|D(r)|}\sum_{b_i \in D(r)}\frac{1 + \text{cos}\Delta_{i}^{\theta}}{2}\delta_{i}
	\end{equation}
	, where $D(r)$ denotes the set of all object predictions at recall rate $r$ and $\Delta^{\theta}_{i}$ is the difference in yaw angle between estimated and ground-truth orientation for object $i$. If $b_i$ is a 2D false positive, i.e., its 2D intersection-over-union (IoU) with ground truth is smaller than a threshold (0.5 or 0.7), $\delta_i = 0$. Note that $AOS$ itself builds on the official average precision metric $AP_{2D}$ and is upper-bounded by $AP_{2D}$. $AOS = \text{1}$ if both the object detection and the orientation estimations are perfect.  
	
	Based on \emph{Minimal Matching Distance} (MMD)~\cite{yuan2018pcn, yu2021pointr}, we propose a new metric $AP_{MMD}$ in the same manner as
	\begin{equation}
	AP_{MMD} = \frac{1}{11}\sum_{r\in\{0, 0.1, \dots, 1\}}\text{max}_{\tilde{r}:\tilde{r}\geq r}s_{MMD}(\tilde{r})
	\end{equation}
	where $r$ is the same detection recall and $s_{MMD}(r) \in [0, 1]$ is the MMD similarity (MMDS) at recall level $r$. MMDS is defined as 
	
	\begin{equation}
	\resizebox{0.48\textwidth}{!}{	
		$
		s_{MMD}(r) = \frac{1}{|D(r)|}\sum_{b_i \in D(r)}[(\gamma - MMD(\compi) * \frac{1}{\gamma}]\delta_i^{MMD},
		$
	}	
	\end{equation}
	
	where $\delta_i^{MMD}$ is a indicator and $MMD(\compi)$ denotes the MMD of prediction $i$ which measures the quality of the predicted surface. If the $i$-th prediction is a false postive or $MMD(\compi) > \gamma$, $\delta_i^{MMD} = 0$. $AP_{MMD}$ is thus also upper-bounded by the official $AP_{2D}$. $AP_{MMD} = AP_{2D}$ if and only if $MMD(\compi) = 0$ for all predictions. In experiments we set $\gamma$ as 0.05.
	
	Since instance-level ground truth shape is not available in KITTI, $MMD(\compi)$ is implemented as category-level similarity similar to~\cite{yuan2018pcn, yu2021pointr}. For a predicted instance PC $\compi$, it is defined as the minimal $L_2$ Chamfer distance between it and a collection of template PCs in ShapeNet~\cite{chang2015shapenet} that has the same class label. It is formally expressed as
	$
	MMD(\compi) = \min_{\mathcal{G} \in SN} d_{CD}(\compi, \mathcal{G}),
	$
	where $SN$ stands for the set of ShapeNet template PCs and $d_{CD}(\compi, \mathcal{G})$ is defined as
	\begin{equation}
	d_{CD}(\compi, \mathcal{G}) = \frac{1}{|\compi|}\sum_{p\in\compi}\min_{g \in \mathcal{G}}||p - g|| + \frac{1}{|\mathcal{G}|}\sum_{g\in\mathcal{G}}\min_{p \in \compi}||g - p||.
	\end{equation}
	During the evaluation, we downloaded 250 ground truth car PCs that were used in \cite{yuan2018pcn, yu2021pointr} for consistency. 
	
	The evaluation of $AP_{MMD}$ considers false negatives. For completeness, we also design \emph{True Positive Minimal Matching Distance} (MMDTP) to evaluate MMD for the true positive predictions similar to~\cite{caesar2020nuscenes}. We define MMDTP@$\beta$ as the average MMD of the predicted objects that have 3D IoU $>$ $\beta$ with at least one ground truth object,
	\begin{equation}
	\resizebox{0.48\textwidth}{!}{	
		$
		MMDTP@\beta = \frac{1}{\sum_{i=1}^{N}TP(i)}\sum_{i=1}^{N}TP(i)*MMD(\compi),
		$
	}	
	\end{equation}
	where $TP(i)$ is 1 if $b_i$ is a true positive and 0 otherwise as 
	\begin{align*}
	TP(i) = 
	\begin{cases}
	1, 
	&\mbox{if } IoU_{3D}(b_i, gt) > \beta, \exists gt\\
	0, 
	& 
	\mbox{else}.
	\end{cases}
	\end{align*}
	
	\section{Experiments}
	\label{sec:experiments}
	We first introduce the used benchmark dataset and the evaluation metrics, followed by a system-level comparison between our S-3D-RCNN with other approaches in terms of outdoor 3D scene understanding capabilities. We further present a module-level comparison to demonstrate the effectiveness of Ego-Net++. Finally, we conduct an ablation study on key design factors and hyper-parameters in EgoNet++. For more training and implementation details, please refer to our supplementary material.
	\subsection{Experimental settings}
	\noindent\textbf{Dataset and evaluation metrics.} We employ the KITTI object detection benchmark~\cite{geiger2012we} that contains stereo RGB images captured in outdoor scenes. The dataset is split into 7,481 training images and 7,518 testing images. The training images are further split into the \emph{train} split and the \emph{val} split containing 3,712 and 3,769 images respectively. For consistency with prior studies, we use the \emph{train} split for training and report results on the \emph{val} split and the testing images. We use the official average precision metrics as well as our newly designed $AP_{MMD}$ and MMDTP. As defined in Eq.~\ref{eq:def_aos} in the main text, $AOS$ is used to assess the system performance for joint object detection and orientation estimation. \emph{3D Average Precision} ($AP_{3D}$) measures precisions at the same recall values where a true positive prediction has 3D IoU $>$ 0.7 with the ground truth one. \emph{BEV Average Precision} ($AP_{BEV}$) instead uses 2D IoU $>$ 0.7 as the criterion for true positives where the 3D bounding boxes are projected to the ground plane. Each ground truth label is assigned a difficulty level (easy, moderate, or hard) depending on its 2D bounding box height, occlusion level, and truncation level.
	
	\begin{table*}[t]
		\footnotesize
		\begin{center}
			\caption{System-level evaluation by comparing Average Orientation Similarity ($AOS$) with previous learning-based methods on the KITTI test set for the car category. Without using LiDAR data~\cite{ku2019monocular} during training (indicated by +LiDAR) or temporal information~\cite{kinematic-3d}, a monocular system using Ego-Net out-performs previous image-based methods. A stereo system using Ego-Net++ shows further performance improvements.}			
			\resizebox{0.8\textwidth}{!}{			
				\begin{tabular}{|c|c|c|c|c|c|c|c|}
					\hline
					Method &  Modality & Number of Viewpoints & Easy & Moderate & Hard & Average\\
					\hline
					
					Mono3D~\cite{monocular3d} &RGB & Monocular &91.01&86.62 &76.84 &84.82\\
					
					Deep3DBox~\cite{mousavian20173d} & RGB & Monocular&92.90&88.75&76.76 &86.14\\		
					
					ML-Fusion~\cite{xu2018multi} &RGB &Monocular &90.35&87.03 & 76.37 &84.58\\			
					
					FQNet~\cite{liu2019deep} &RGB &Monocular&92.58 &88.72 &76.85 &86.05\\
					
					GS3D~\cite{li2019gs3d} &RGB &Monocular&85.79  &75.63  & 61.85 &74.42\\
					
					MonoPSR~\cite{ku2019monocular} &RGB + LiDAR &Monocular&93.29 &	87.45 &	72.26 &84.33\\
					
					M3D-RPN~\cite{brazil2019m3d} &RGB &Monocular&88.38 &82.81& 67.08 &79.42\\
					
					MonoPair~\cite{chen2020monopair} &RGB &Monocular&91.65 &86.11 &76.45 &84.74\\	
					
					Disp R-CNN~\cite{sun2020disp} &RGB + LiDAR &Stereo &93.02 &	81.70 &	67.16 &80.63\\				
					
					DSGN~\cite{chen2020dsgn} &RGB &Stereo&95.42&86.03& 78.27 &86.57\\	
					
					D4LCN~\cite{Ding_2020_CVPR} &RGB &Monocular&90.01&82.08& 63.98 &78.69\\		
					
					RAR-Net~\cite{liu2020reinforced}&RGB &Monocular&88.40&82.63 &66.90 &79.31\\	
					
					RTM3D~\cite{RTM3D} &RGB &Monocular&91.75&86.73& 77.18 &85.22\\								
					
					Kinematic3D~\cite{kinematic-3d} &RGB &Monocular&58.33 &	45.50 &	34.81 &46.21\\
					\hline
					
					Ours (Ego-Net)~\cite{Li_2021_CVPR}&RGB &Monocular&\textbf{96.11}&\textbf{91.23}& \textbf{80.96} &\textbf{89.43} \\
					
					Ours (Ego-Net++) &RGB &Stereo&\textbf{96.38}&\textbf{93.58}& \textbf{88.00} &\textbf{92.65} \\				
					\hline
				\end{tabular}
				\label{tab:orientation}
			}
		\end{center}
		
	\end{table*}
	
	\begin{table*}
		\footnotesize
		\begin{center}
			\caption{Comparison of $AOS$ with other methods on the KITTI test set for the pedestrian category. Without using LiDAR data~\cite{lang2019pointpillars}, a stereo system using Ego-Net++ significantly outperforms previous studies. The proposals are the same as GUPNet~\cite{lu2021geometry}.}
			\label{tab:orientation_ped}			 		
			\resizebox{0.8\textwidth}{!}{
				\begin{tabular}{|c|c|c|c|c|c|c|c|}
					\hline
					Method &  Modality&  Number of Viewpoints & Easy & Moderate & Hard & Average\\
					\hline
					PointPillars~\cite{lang2019pointpillars} & LiDAR & N/A & 57.47 & 48.05 &	45.40 	& 50.31 \\
					PointRCNN~\cite{shi2019pointrcnn}  & LiDAR & N/A & 57.19 &	47.33 &	44.31 & 49.61 \\
					M3D-RPN~\cite{brazil2019m3d} &RGB & Monocular&44.33 &	31.88 &	28.55 &34.92\\
					TANet~\cite{liu2020tanet}  & LiDAR & N/A & 42.54 &	36.21 &	34.39 & 37.71\\
					Disp R-CNN~\cite{sun2020disp}&RGB + LiDAR & Stereo &63.16 &	45.66 &	41.14 &50.00\\	
					D4LCN~\cite{Ding_2020_CVPR} &RGB & Monocular &46.73 &	33.62 &	28.71 &36.35\\						
					DSGN~\cite{chen2020dsgn}&RGB & Stereo &31.21 &	24.32 &	23.09 &26.21\\	
					YOLOStereo~\cite{liu2021yolostereo3d}  & RGB & Stereo& 48.99 &	35.62 &	31.58 & 38.73 \\
					MonoEF~\cite{Zhou_2021_CVPR} & RGB & Monocular & 47.45 &	34.63 &	31.01 & 37.70 \\
					CaDDN~\cite{reading2021categorical} & RGB & Monocular& 24.45 &	17.13 &	15.79 & 19.12 \\
					LIGA-Stereo~\cite{guo2021liga}& RGB + LiDAR & Stereo &53.16 &	40.98 &	38.12 & 44.09 \\
					GUPNet~\cite{lu2021geometry}& RGB & Monocular& 68.93 &	50.74 &	44.01 & 54.56 \\
					MonoCon~\cite{monocon} & RGB & Monocular & 52.16 &	38.67 &	33.15 & 41.33 \\
					\hline
					Ours &RGB & Stereo &\textbf{73.14}&\textbf{53.92}& \textbf{46.93} &\textbf{57.98} \\
					\hline
				\end{tabular}
				
			}
		\end{center}	
		
	\end{table*}
	
	\begin{figure*}[t]
		\begin{center}
			\centering
			\includegraphics[width=1.\linewidth, trim=0cm 0cm 0cm 0cm]{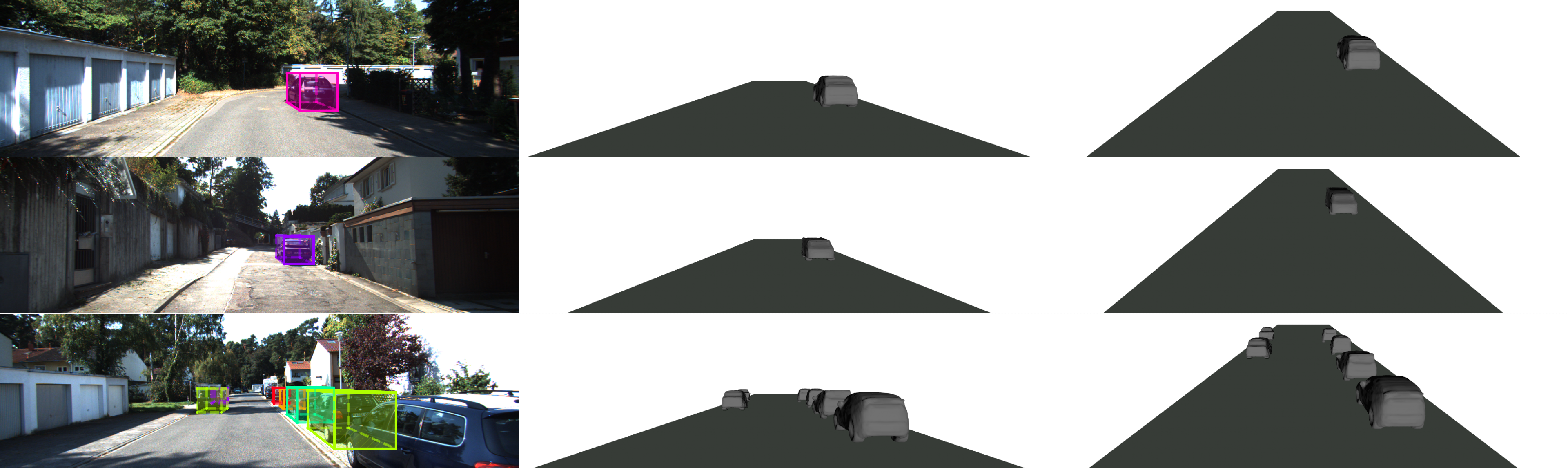}
			\captionof{figure}{Qualitative results of S-3D-RCNN on KITTI \emph{val} set. Left: input left images and 3D proposals. Middle: rendered objects in the original camera view. Right: rendered objects in a different view (camera 2 meters higher with 15$^\circ$ pitch angle).}
			\label{qualitative_results}
		\end{center}
	\end{figure*} 
	
	\begin{table*}[t]
		\footnotesize
		\centering
		\caption{A comparison of image-based outdoor 3D scene understanding performance on the KITTI \emph{val} set for the Car class. When evaluating MMDTP, the predictions are grouped according to the predicted depth.}	
		\resizebox{\textwidth}{!}{
			\begin{tabular}{|c|c|c|c|c|c|c|c|}
				\hline
				
				\multicolumn{8}{|c|}{Qualitative comparison of perception capability} \\
				\hline
				
				\multirow{2}{*}{Method} & \multirow{2}{*}{Modality} & \multirow{2}{*}{Location} & \multirow{2}{*}{Dimension} & \multirow{2}{*}{Orientation} & \multicolumn{3}{c|}{Shape Representation}  \\ 
				& & & & & Instance point cloud& Mesh& Implicit shape\\	 
				\hline
				3D-RCNN~\cite{kundu20183d}  & Monocular & \checkmark& \checkmark& \checkmark&  &\checkmark &\\
				\hline 
				DSGN~\cite{chen2020dsgn}   & Stereo & \checkmark& \checkmark& \checkmark& \checkmark~(Partial) & &\\
				\hline
				Disp-RCNN~\cite{sun2020disp}  & Stereo & \checkmark& \checkmark& \checkmark& \checkmark~(Partial) & & \\		 
				\hline		 
				S-3D-RCNN (Ours)  & Stereo & \checkmark& \checkmark& \checkmark& \checkmark~(Full) &\checkmark & \checkmark\\
				\hline
				
				\multicolumn{8}{|c|}{Quantitative comparison of 3D object detection performance} \\
				\hline
				\multicolumn{2}{|c|}{\multirow{2}{*}{Method}} &  \multicolumn{3}{c}{$AP_{3D}@R_{11}\uparrow$} & \multicolumn{3}{|c|}{$AP_{BEV}@R_{11}\uparrow$} \\ 
				
				\multicolumn{2}{|c|}{~} & Easy & Moderate & Hard & Easy & Moderate & Hard\\ \hline 
				\multicolumn{2}{|c|}{3D-RCNN~\cite{kundu20183d}} &N/A&N/A&N/A&N/A&N/A&N/A\\ 
				\hline
				\multicolumn{2}{|c|}{DSGN~\cite{chen2020dsgn}} & 72.31 & 54.27 & 47.71 & 83.24 & 63.91 & 57.83\\
				\hline
				\multicolumn{2}{|c|}{Disp-RCNN~\cite{sun2020disp}} & 64.29 & 47.73 & 40.11 & 77.63 & 64.38 & 50.68 \\
				\hline
				\multicolumn{2}{|c|}{S-3D-RCNN (Ours)} & {\bf 77.04}  & {\bf 62.98}  & {\bf 56.56} & {\bf 87.37}  & {\bf72.87}  & {\bf66.62}\\	
				\hline	
				
				\multicolumn{8}{|c|}{Quantitative comparison of shape reconstruction performance with MMDTP@0.5$\downarrow$}  \\
				\hline
				\multicolumn{2}{|c|}{Depth range}  &(0,10m] &(10,20m]&(20,30m]&(30,40m]&(40,50m]&(50,60m] \\	
				\hline
				\multicolumn{2}{|c|}{DSGN~\cite{chen2020dsgn}}  & 0.036&
				0.039&
				0.041&
				0.044&
				0.048&
				0.051\\ 
				\hline		
				\multicolumn{2}{|c|}{Disp-RCNN~\cite{sun2020disp}} & 0.034&
				0.037&
				0.039&
				0.038&
				0.041&
				0.034\\
				\hline
				\multicolumn{2}{|c|}{Disp-RCNN~\cite{sun2020disp}+ $\mathcal{E}$}  
				&{\bf0.0051}&
				{\bf0.0071}&
				{\bf0.0074}&
				{\bf0.0065}&
				{\bf0.0050}&
				{\bf0.0035}\\
				\hline
			\end{tabular}
		}
		\label{table:comparison}
	\end{table*}
	
	\subsection{System-level comparison with previous studies}
	Ego-Net++ can be combined with a 2D proposal model to build a strong system for object orientation/rigid shape recovery.
	\subsubsection{Joint object detection and orientation estimation performance}
	Ego-Net can be used with a 2D vehicle detection model to form a joint vehicle detection and orientation estimation system, whose performance is measured by $AOS$. Per the proposals used in~\cite{Li_2021_CVPR}, Tab.~\ref{tab:orientation} compares the $AOS$ of our system using Ego-Net with other approaches on the KITTI test set for the car category. Among the single-view image-based approaches, our system outperforms others by a clear margin. Our approach using a single image outperforms Kinematic3D~\cite{kinematic-3d} which exploits temporal information using RGB video.
	
	In addition to monocular orientation estimation, we show system performance with our $\mathcal{E}$ that can exploit two-view information in Tab.~\ref{tab:orientation} and Tab.~\ref{tab:orientation_ped} for the car and pedestrian class respectively. Our S-3D-RCNN using $\mathcal{E}$ achieves improved vehicle orientation estimation performance. For performance comparison of pedestrian orientation estimation, we use the same proposals in~\cite{lu2021geometry}. Using our $\mathcal{E}$ consistently outperforms the proposal model, and the system performance based on RGB images even surpasses some LiDAR-based approaches~\cite{lang2019pointpillars, shi2019pointrcnn} that have more accurate depth measurements. This result indicates that the LiDAR point cloud is not discriminative for determining the accurate orientation of distant non-rigid objects due to a lack of fine-grained visual information. In contrast, our image-based approach effectively addresses the limitations of LiDAR sensors and can complement them in this scenario.
	
	\subsubsection{Comparison of 3D scene understanding capability}
	To our knowledge, S-3D-RCNN is the first model that jointly performs accurate 3DOD and implicit rigid shape estimation for outdoor objects with stereo cameras. Tab.~\ref{table:comparison} presents a summary and comparison of perception capability with previous image-based outdoor 3D scene understanding approaches. Qualitatively, S-3D-RCNN is the only method that can utilize stereo geometry as well as predict implicit shape representations. Compared to the monocular method 3D-RCNN~\cite{kundu20183d} that uses template meshes with fixed topology, our framework can produce meshes in a resolution-agnostic way and can provide an accurate estimation of object locations by exploiting two-view geometry. We show qualitative results in Fig.~\ref{qualitative_results} where the predicted implicit shapes are decoded to meshes. Our approach shows accurate localization performance along with plausible shape predictions, which opens up new opportunities for outdoor augmented reality.  
	
	\subsection{Module-level comparison with previous studies}
	Here we demonstrate the effectiveness of EgoNet++ as a module. Based on the designed IGRs and progressive mappings in this study, we show how using Ego-Net++ can contribute to improved orientation/shape estimation performance for state-of-the-art 3D scene understanding approaches.
	\subsubsection{Comparison of orientation estimation performance}
	To assess if Ego-Net can help improve the pose estimation accuracy of other 3DOD systems, we download proposals from other open-source implementations and use Ego-Net for orientation refinement. The result is summarized in Tab.~\ref{tab:module} for the car category. While $AOS$ depends on the detection performance of these methods, using Ego-Net consistently improves the pose estimation accuracy of these approaches. This indicates that Ego-Net is robust despite the performance of a vehicle detector varying with different recall levels. We also compare with OCM3D~\cite{peng2021ocm3d} using the same proposals, and higher $AOS$ further validates the effectiveness of our proposed IGRs for orientation estimation. 
	
\begin{figure}
		\begin{center}
			\includegraphics[width=0.6\linewidth, trim=0cm 0cm 0cm 0cm]{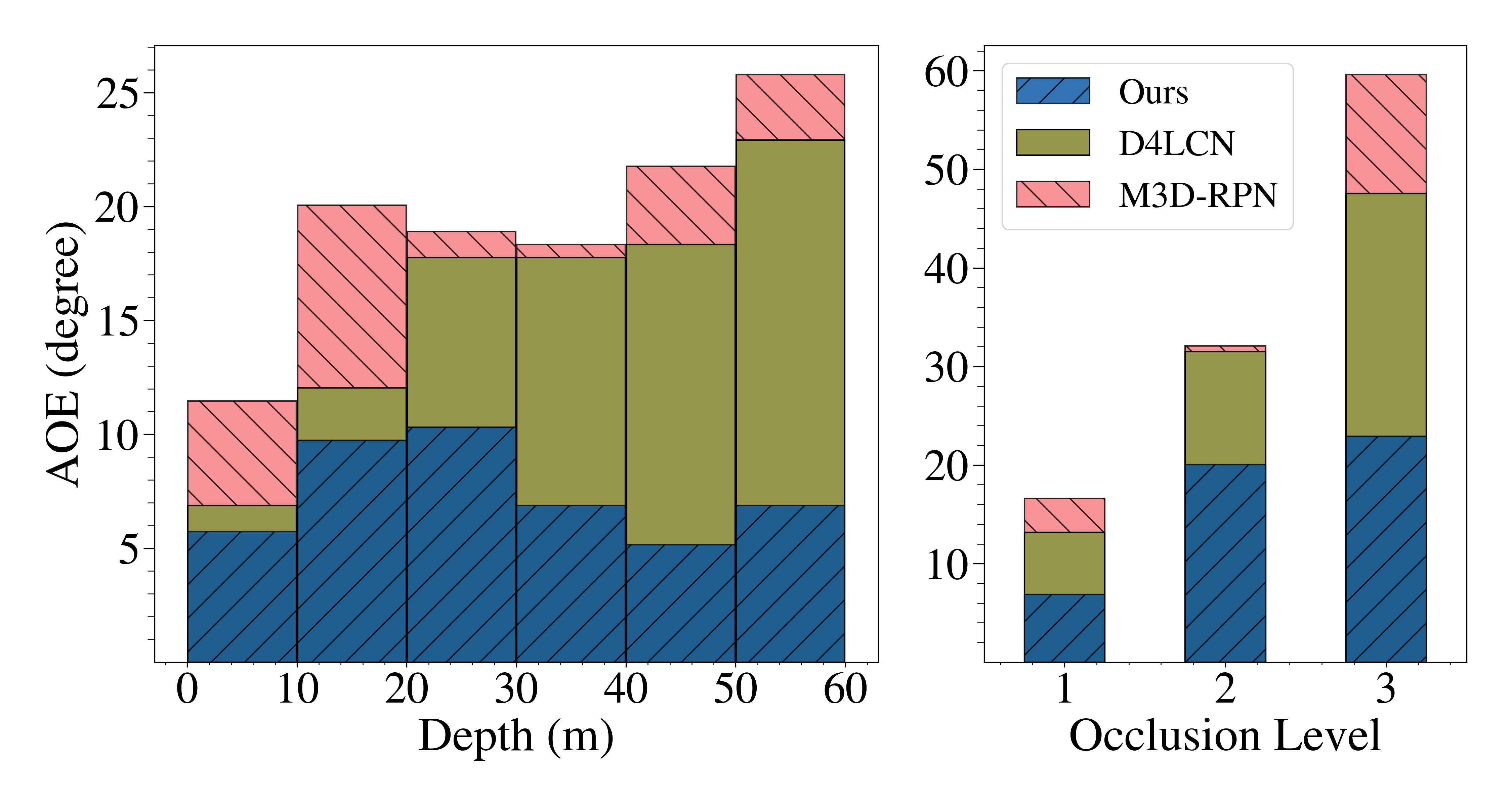}
		\end{center}
		\caption{Average orientation error (AOE) on KITTI \emph{val} split in different depth ranges and occlusion levels. Our approach is robust to distant and partially occluded instances.}
		\label{fig:aoe}
	\end{figure}

	For true positive predictions we plot the distribution of orientation estimation error versus different depth ranges and occlusion levels in Fig.~\ref{fig:aoe}. The error in each bin is the averaged orientation estimation error for those instances that fall into it. While the performance of M3D-RPN~\cite{brazil2019m3d} and D4LCN~\cite{Ding_2020_CVPR} degrades significantly for distant and occluded cars, the errors of our approach increase gracefully. We believe that explicitly learning the object parts makes our model more robust to occlusion as the visible parts can provide richer information for pose estimation.
	
	\begin{table}[h]
		\centering
		\small
		\captionof{table}{$AOS$ evaluation on KITTI validation set. After employing Ego-Net, the vehicle pose estimation accuracy of other 3DOD systems can be improved. The space for $AOS$ improvement is upper-bounded by $AP_{2D}$.}	
		\begin{tabular}{|l|c|c|c|}
			\hline
			\multirow{3}{*}{Method} & \multicolumn{3}{|c|}{$AOS\&AP_{2D}$,~~$AOS \le AP_{2D}$}\\ \cline{2-4}
			&  Easy & Moderate & Hard \\ \cline{2-4}
			& \multicolumn{3}{|c|}{$AP_{2D}$}\\ 
			\hline
			
			M3D-RPN~\cite{brazil2019m3d}  &  90.28 & 83.75  & 67.72 \\
			
			D4LCN~\cite{Ding_2020_CVPR}  & 92.80 & 84.43  & 67.89 \\
			\hline
			Method	& \multicolumn{3}{c|}{$AOS$}\\ 
			\hline
			
			M3D-RPN~\cite{brazil2019m3d}  &  88.79 & 81.25  & 65.37 \\
			
			M3D-RPN + Ego-Net  &  90.20 &83.60  & 67.53 \\
			
			D4LCN~\cite{Ding_2020_CVPR}  & 91.74  &82.96  &66.45\\
			
			D4LCN + OCM3D~\cite{peng2021ocm3d} &  92.12 &83.27 &66.81 \\	
			
			D4LCN + Ego-Net  &  92.62 &84.25  &67.60 \\	
			\hline	
		\end{tabular}
		\label{tab:module}
	\end{table}
	
	\begin{table}[h]
		\centering
		\small
		\captionof{table}{$AP_{BEV}$ evaluated on KITTI validation set. Ego-Net can correct the erroneous pose predictions from~\cite{Ding_2020_CVPR} as shown in Fig.~\ref{fig:quali_comp}.}	
		\begin{tabular}{|l|c|c|c|}
			\hline
			\multirow{2}{*}{Method} & \multicolumn{3}{c|}{$AP_{BEV}$}\\ \cline{2-4}
			&  Easy & Moderate & Hard \\ 
			\hline
			
			ROI-10D~\cite{manhardt2019roi}  &  14.04 &3.69  &3.56 \\
			
			Mono3D++~\cite{he2019mono3d++}  &  16.70 &11.50  &10.10 \\
			
			MonoDIS~\cite{disentangling}  &  24.26 &18.43  &16.95 \\
			
			D4LCN~\cite{Ding_2020_CVPR}   &31.53 &22.58  &17.87 \\
			
			D4LCN + Ego-Net (Ours)  &  \textbf{33.60} &\textbf{25.38} &\textbf{22.80} \\	
			\hline
		\end{tabular}
		\label{tab:bev}
	\end{table}
	
	\begin{figure}[!htb]
		\centering
		\begin{minipage}{.48\textwidth}
			\centering
			\includegraphics[width=\linewidth]{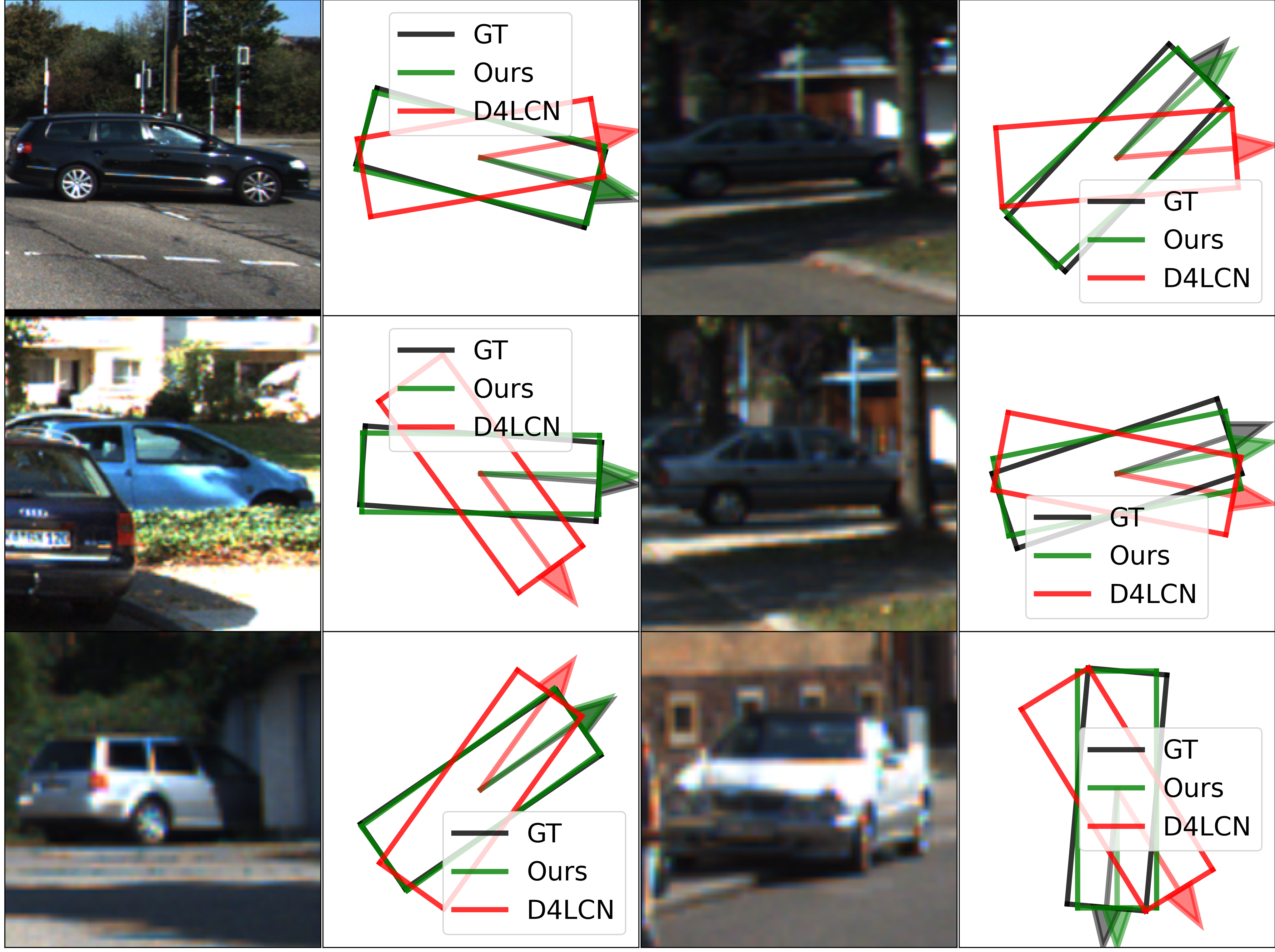}
			\captionof{figure}{Detected instances on KITTI \emph{val} split along with the comparison of the predicted vehicle orientations in bird's eye view.}
			\label{fig:quali_comp}
		\end{minipage}%
		\hfill
		\begin{minipage}{.48\textwidth}
			\centering
			\includegraphics[width=\linewidth]{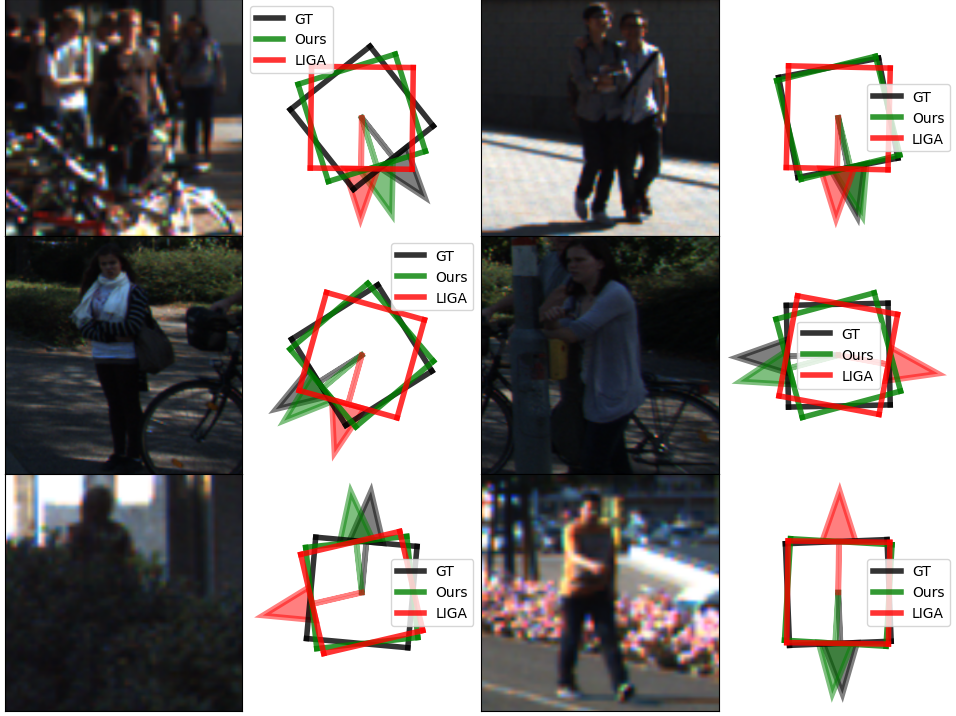}
			\captionof{figure}{Similar comparison as Fig.~\ref{fig:quali_comp} showing pedestrian orientation predictions using two-view images as inputs.}
			\label{pedes_quali}
		\end{minipage}
	\end{figure}
	
	We visualize qualitative comparison with D4LCN~\cite{Ding_2020_CVPR} in Fig.~\ref{fig:quali_comp} in Bird's eye view (BEV). The orientation predictions are shown in BEV and the arrows point to the heading direction of those vehicles. From the local appearance, it can be hard to tell whether certain cars head towards or away from the camera, as validated by the erroneous predictions of D4LCN~\cite{Ding_2020_CVPR} since it regresses pose from local features. In comparison, our approach gives accurate egocentric pose predictions for these instances and others that are partially occluded. Quantitative comparison with several 3D object detection systems~on the KITTI \emph{val} split is shown in Tab.~\ref{tab:bev}. Note that utilizing Ego-Net can correct wrong pose predictions especially for difficult instances, which leads to improved 3D IoU and results in significantly improved $AP_{BEV}$.
	
	\begin{table}
		\centering
		\caption{Quantitative comparison for orientation predictions for pedestrians on the KITTI \emph{val} set using the same proposals as~\cite{guo2021liga}. 40 recall values are used for consistency.}	
		\begin{tabular}{|l|c|c|c|}
			\hline
			\multirow{2}{*}{Method} & \multicolumn{3}{c|}{$AP_{3D}@R_{40}$}\\ \cline{2-4}
			&  Easy & Moderate & Hard \\ 
			\hline
			LIGA~\cite{guo2021liga}   &45.54 &37.80 &32.09 \\	
			LIGA + Ego-Net++ (Ours)  &  \textbf{50.25} &\textbf{44.48} &\textbf{38.63} \\	
			\hline
			\multirow{2}{*}{Method} & \multicolumn{3}{c|}{$AP_{BEV}@R_{40}$}\\ \cline{2-4}
			&  Easy & Moderate & Hard \\ 
			\hline
			LIGA~\cite{guo2021liga}   &54.59 &47.14 &41.12 \\	
			LIGA + Ego-Net++ (Ours)  &  \textbf{59.55} &\textbf{53.35} &\textbf{47.96} \\
			\hline
			\multirow{2}{*}{Method} & \multicolumn{3}{c|}{$AOS@R_{40}$}\\ \cline{2-4}
			&  Easy & Moderate & Hard \\ 
			\hline
			LIGA~\cite{guo2021liga}   &54.64 &48.22 &43.10 \\	
			LIGA + Ego-Net++ (Ours)  &  \textbf{73.56} &\textbf{69.28} &\textbf{62.92} \\
			\hline
		\end{tabular}		
		\label{pedes_quanti}
	\end{table}
	
	Apart from using Ego-Net for monocular detectors, we provide extended studies of using Ego-Net++ for a state-of-the-art stereo detector. Tab.~\ref{pedes_quanti} shows a comparison for the non-rigid pedestrian class where the same proposals are used as~\cite{guo2021liga}. Thanks to the effectiveness of PPC, our $\mathcal{E}$ significantly improves the 3DOD performance for these non-rigid objects and some examples are shown in Fig.~\ref{pedes_quali}. This improvement shows the IGRs designed in this study are robust despite the training data for the pedestrian class being much fewer.
	
	\begin{figure}[!htb]
		\small
		\centering
		\begin{minipage}{.48\textwidth}
			\centering
			\includegraphics[width=\linewidth]{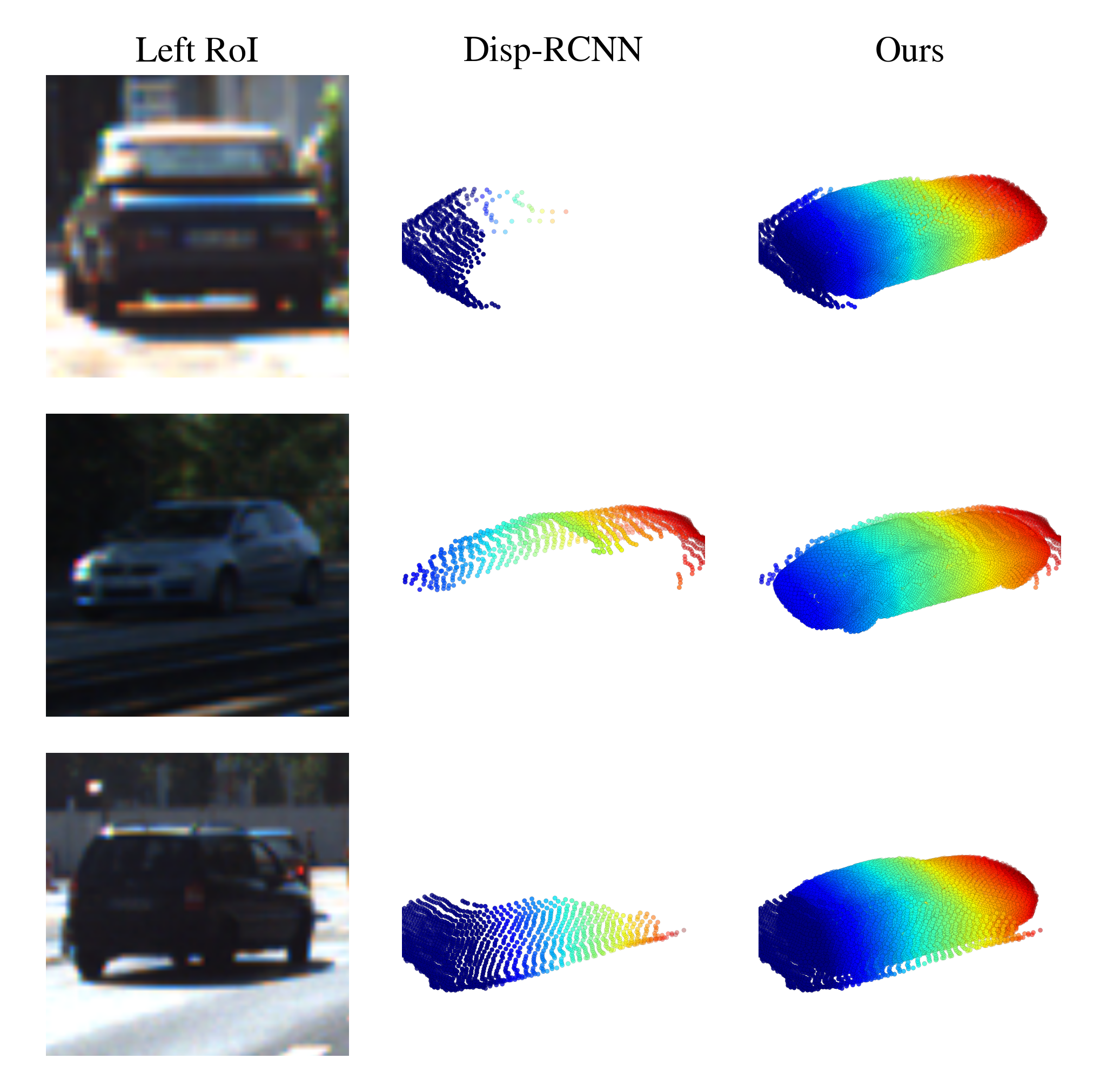}
			\captionof{figure}{Qualitative comparison for instances on the KITTI \emph{val} split. Left: instance RoIs in the left image. Middle: instance point cloud predictions from Disp R-CNN~\cite{sun2020disp}. Right: Our predictions $\compi$ after the hallucination module.}
			\label{qualitative_comp}
		\end{minipage}%
		\hfill
		\begin{minipage}{.48\textwidth}
			\centering
			\includegraphics[width=\linewidth]{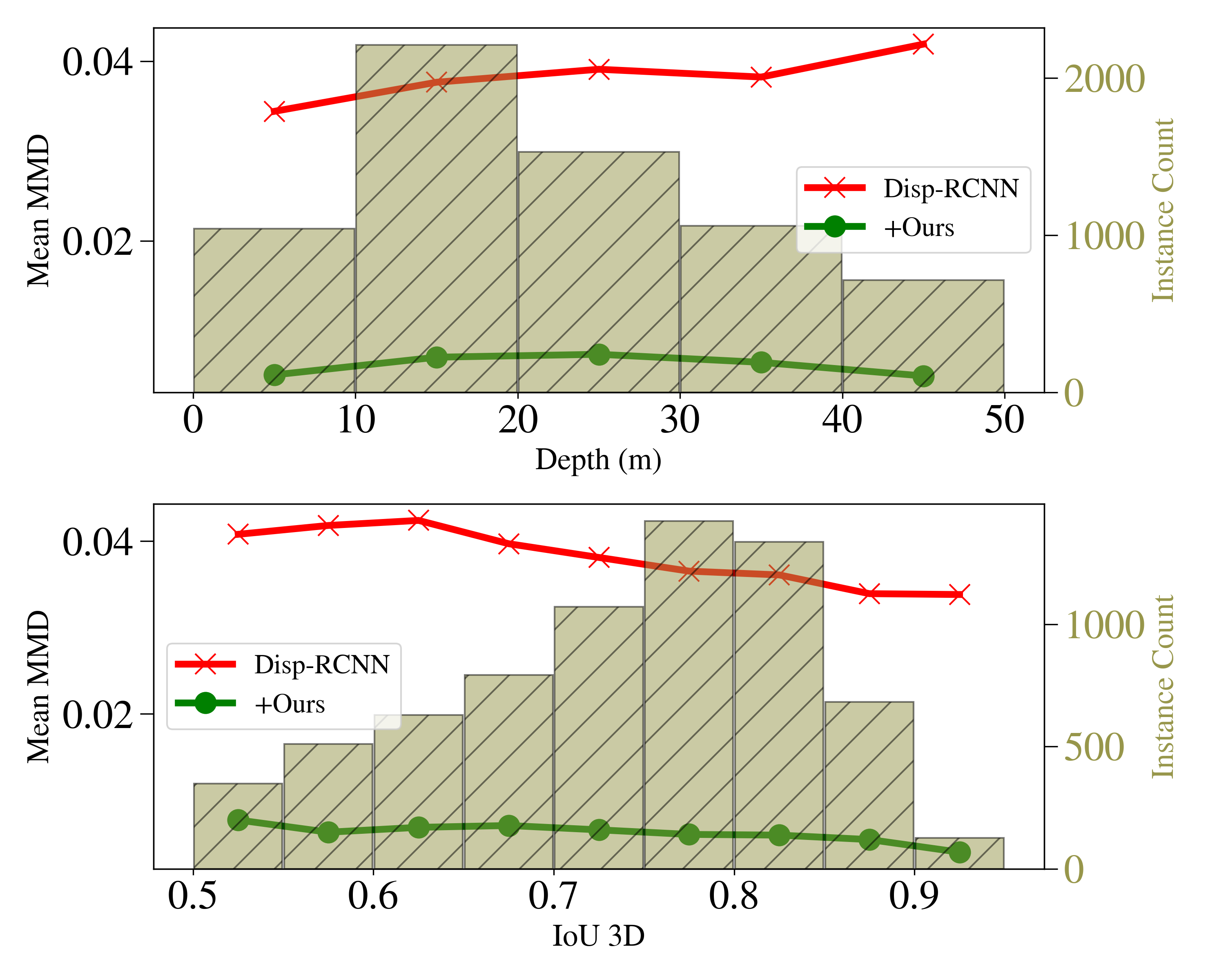}
			\captionof{figure}{The bar plots show the distribution of the predicted bounding boxes used in MMDTP evaluation. The average MMD for each bin is shown in the line plots. Using $\mathcal{E}$ consistently improves performance for objects in all bins.}
			\label{fig:mmdtp}
		\end{minipage}
	\end{figure}
	
	\subsubsection{Comparison of shape estimation performance}
	Here we present a quantitative comparison of rigid shape estimation quality using our new metric MMDTP. We compare with \cite{sun2020disp} due to its availability of the official implementation. We downloaded 13,575 predicted objects from its website in which 9,488 instances have 3D IoU large enough to compute MMDTP@0.5 shown in~Tab.~\ref{table:comparison}. Our $\mathcal{E}$ can produce a complete shape description of a detected instance thanks to our \emph{visible surface representation} and explicit modeling of the unseen surface hallucination problem. This leads to a significant improvement in MMDTP.  Fig.~\ref{qualitative_comp} shows a qualitative comparison of the predicted instance shapes as PCs. Note that the visible portion of the instances varies from one to another, but our $\mathcal{E}$ can reliably infer the invisible surface. Fig.~\ref{fig:mmdtp} shows the relationship between MMDTP with factors such as object depth and bounding box quality. More distant objects and less accurate 3D bounding boxes suffer from larger MMD for~\cite{sun2020disp} due to fewer visible points and larger alignment errors. Note that our approach consistently improves~\cite{sun2020disp} across different depth ranges and box proposal quality levels. 
	
	\begin{table}
		\footnotesize
		\centering
		\caption{Quantitative comparison using our introduced new metric $AP_{MMD}$ on KITTI \emph{val} split.}	
		\begin{tabular}{|c|c|c|c|}
			\hline
			\multirow{2}{*}{Method} & \multicolumn{3}{c|}{$AP_{MMD}@R_{11}$ (2D IoU $>$ 0.7)}\\ \cline{2-4}
			&Easy & Moderate & Hard \\ \hline 
			Disp-RCNN \cite{sun2020disp} &  37.50 & 30.93& 25.29 \\
			Disp-RCNN + $\mathcal{E}$ (Ours) & 89.10& 83.71& 72.26 \\
			\hline
		\end{tabular}
		\label{table:AP_MMD}
	\end{table}
	
\begin{figure*}[!htb]
		\centering
		\begin{minipage}[t]{0.45\textwidth}
			\centering
			\includegraphics[width=\linewidth]{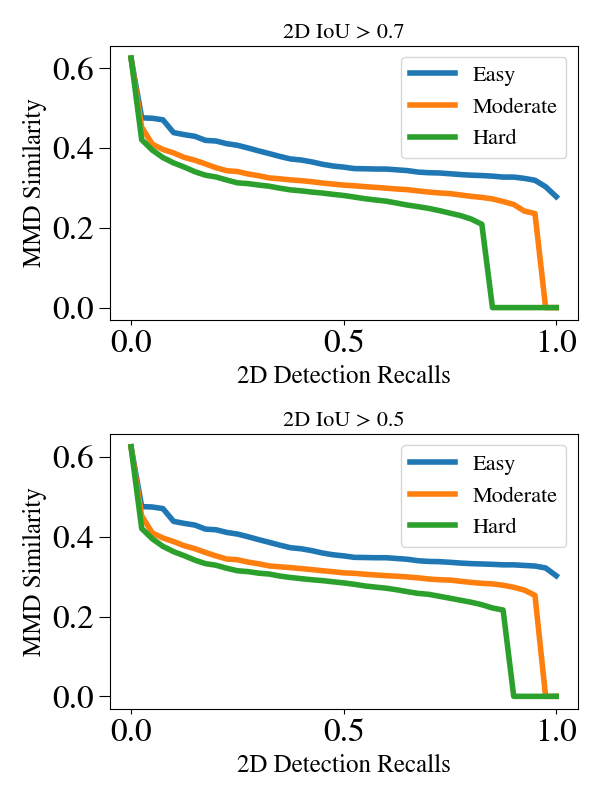}
			\captionof{figure}{MMD similarity of Disp R-CNN~\cite{sun2020disp} used for computing the newly introduced metric $AP_{MMD}$. Top: 2D IoU $>$0.7 is used to determine a true positive. Bottom: 2D IoU $>$0.5 is used instead.}
			\label{AP_MMD_disp}
		\end{minipage}
		\hfill
		\begin{minipage}[t]{0.45\textwidth}
			\centering
			\includegraphics[width=\linewidth]{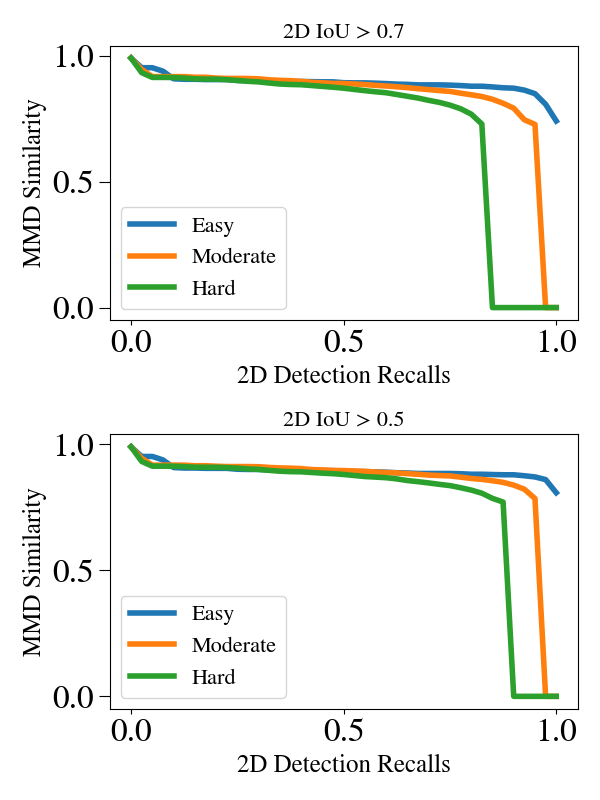}
			\captionof{figure}{The same plot as Fig.~\ref{AP_MMD_disp} for Disp R-CNN~\cite{sun2020disp}+$\mathcal{E}$. Using our Ego-Net++ significantly improves $AP_{MMD}$.}
			\label{AP_MMD}
		\end{minipage}
	\end{figure*}

	Per our definition of $AP_{MMD}$, the $s_{MMD}$ at different recall values are shown in Fig.~\ref{AP_MMD_disp} for the predictions of Disp-RCNN. In contrast, the performance of using our $\mathcal{E}$ for the same proposals is shown in Fig.~\ref{AP_MMD}. The detailed quantitative results are shown in Tab.~\ref{table:AP_MMD}. The $AP_{2D}$ (IoU $>$ 0.7), i.e., the upper bound for $AP_{MMD}$, is 99.16, 93.22, 81.28 for easy, moderate, and hard categories respectively. Note our $\mathcal{E}$ has contributed a significant improvement compared with~\cite{sun2020disp}. This indicates our approach has greatly complemented existing 3DOD approaches with the ability to describe outdoor object surface geometry within the 3D bounding boxes.
	
	\begin{table}[h!]
		\centering
		\caption{Module-level evaluation assuming perfect object detection ($AP_{2D}=100.00$) on the KITTI \emph{val} split. B:~baseline using direct regression. +IGR:~adding intermediate geometric representations.}	
		\begin{tabular}{|l|c|c|c|}
			\hline
			\multirow{2}{*}{Method} & \multicolumn{3}{c|}{$AOS$ when $AP_{2D}=100.00$}\\ \cline{2-4}
			&  Easy & Moderate & Hard \\ 
			\hline
			
			B  &  95.22 &92.14  &88.37 \\
			
			Deep3DBox~\cite{mousavian20173d}  &98.48 &95.81  &91.98 \\
			
			B+IGRs (Ego-Net)  &  99.58 &99.06  &96.55\\	
			\hline
		\end{tabular}
		\label{tab:ablation1}
	\end{table} 
	
	\begin{table}[h!]
		\centering
		\captionof{table}{Module-level evaluation assuming perfect object detection on the KITTI \emph{val} split for pedestrians. Ego-Net++ consistently improves over its single-view counterpart.}	
		\begin{tabular}{|l|c|c|c|}
			\hline
			\multirow{2}{*}{Method} & \multicolumn{3}{c|}{$AOS$ when $AP_{2D}=100.00$}\\ \cline{2-4}
			&  Easy & Moderate & Hard \\ 
			\hline
			Ego-Net~\cite{Li_2021_CVPR}  &91.35 &90.93  &89.48 \\
			Ego-Net++  &  91.88 &91.76  &90.23\\
			\hline
		\end{tabular}
		\label{table:egonet}
	\end{table}
	
	\begin{table}[h!]
		\centering
		\captionof{table}{MMDTP evaluated with and without using the $Ha$ module. The same object proposals are used as in Tab.~\ref{table:comparison}.}	
		\begin{tabular}{|c|c|c|c|}
			\hline
			Depth range  &(0,10m] &(10,20m]&(20,30m] \\	
			\hline
			w/o $Ha$ &0.027&0.029&0.032 \\ 
			\hline
			w $Ha$ &0.0051&0.0071&0.0074\\
			\hline
		\end{tabular}
		\label{table:H}
	\end{table}
	
	\subsection{Ablation study}
	\noindent \textbf{Direct regression vs. learning IGRs.} To validate the design of our proposed IGRs for orientation estimation, we compare the pose estimation accuracy of our approach to a baseline that directly regresses pose angles from the instance feature maps. To eliminate the impact of the used object detector, we compare $AOS$ on all annotated vehicle instances in the validation set. This is equivalent to measuring $AOS$ with $AP_{2D}=1$ so that the orientation estimation accuracy becomes the only focus. The comparison is summarized in Tab.~\ref{tab:ablation1}. Note learning IGRs outperforms the baseline by a significant margin. Deep3DBox~\cite{mousavian20173d} is another popular architecture that performs direct angle regression. Our approach outperforms it with the novel design of IGRs. 
	
	\noindent\textbf{Is PPCs better than the single-view representation?} Tab.~\ref{table:egonet} shows the performance comparison between single-view Ego-Net and Ego-Net++ that uses stereo inputs. Using PPCs leads to better performance and validates the improvement of Ego-Net++ over Ego-Net. It also validates the coordinate representations designed in this study can be used easily with multi-view inputs.

	\noindent\textbf{Is $Ha$ useful?} Hereafter we use the same object proposals as used in Tab.~\ref{table:comparison} for consistency. We compare the performance without $Ha$ in Tab.~\ref{table:H}, where the normalized representation $ocsi$ is directly used for evaluation before being passed to $Ha$. The results indicate $Ha$ effectively hallucinates plausible points to provide a complete shape representation. 
	
	\section{Conclusion}
	We propose the first approach for joint stereo 3D object detection and implicit shape reconstruction with a new two-stage model S-3D-RCNN. S-3D-RCNN can (i) perform accurate object localization as well as provide a complete and resolution-agnostic shape description for the detected rigid objects and (ii) produce significantly more accurate orientation predictions. To address the challenging problem of 3D attribute estimation from images, a set of new intermediate geometrical representations are designed and validated. Experiments show that S-3D-RCNN achieves strong image-based 3D scene understanding capability and brings new opportunities for outdoor augmented reality. Our framework can be extended to non-rigid shape estimation if corresponding data is available to train our hallucination module. How to devise an effective training approach to achieve stereo pedestrian reconstruction is an interesting research question which we leave for future work.
	
	
	\section*{Author contributions statement}
	S.Li. conceived the experiment(s),  S.Li. conducted the experiment(s), X.Huang. and Z.Liu. analysed the results.  All authors reviewed the manuscript. 
	
	\section*{Data availability}
	The datasets generated during and/or analyzed during our study are available from the corresponding author on reasonable request.
	

	\section*{Acknowledgement}
	This research was supported by National Natural Science Foundation of China/HKSAR Research Grants Council Joint Research Scheme under Grant N\_HKUST627/20.	
	
	\setcounter{section}{0}
	\clearpage	
	\begin{center}
		\large{\textbf{Supplementary Material}}
	\end{center}	
	This is the supplementary material for \textit{Joint Stereo 3D Object Detection and Implicit Surface Reconstruction}, which contains more details about the model hyper-parameters and the training setting.
	
	\section{Network architecture}
	The detection network used in the experiments is visualized in Fig. 3 of the main text. It has a similar architecture as LIGA~\cite{guo2021liga}. Its detailed hyper-parameters are shown below.
	\begin{figure*}[h!]
		\begin{center}
			\centering
			\includegraphics[width=0.8\linewidth, trim=0cm 0cm 0cm 0cm]{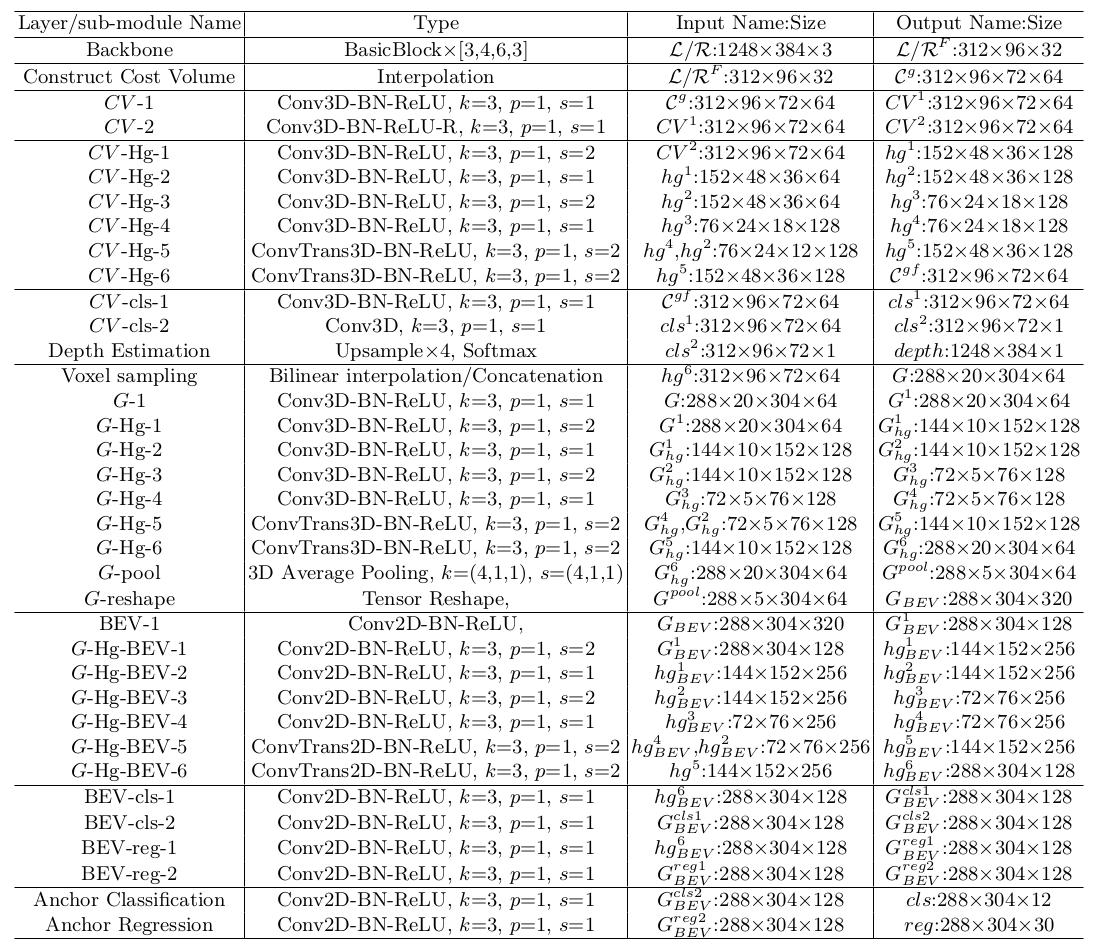}
			\caption{Detailed architecture of the proposal model. Conv3D-BN-
				ReLU is a 3D convolution layer followed by batch normalization and ReLU activation. ConvTrans3D denotes a transposed 3D convolution layer. Conv3D-BN-ReLU-R means the output is added to the input as the residual. k denotes	kernel size. p denotes padding. s denotes stride. d denotes dilation. Hg denotes the hourglass sub-module. BasicBlock denotes a basic block in ResNet.}
			\label{proposal model}
		\end{center}
	\end{figure*}
	
\begin{figure*}[h!]
		\begin{center}
			\centering
			\includegraphics[width=0.8\linewidth, trim=0cm 0cm 0cm 0cm]{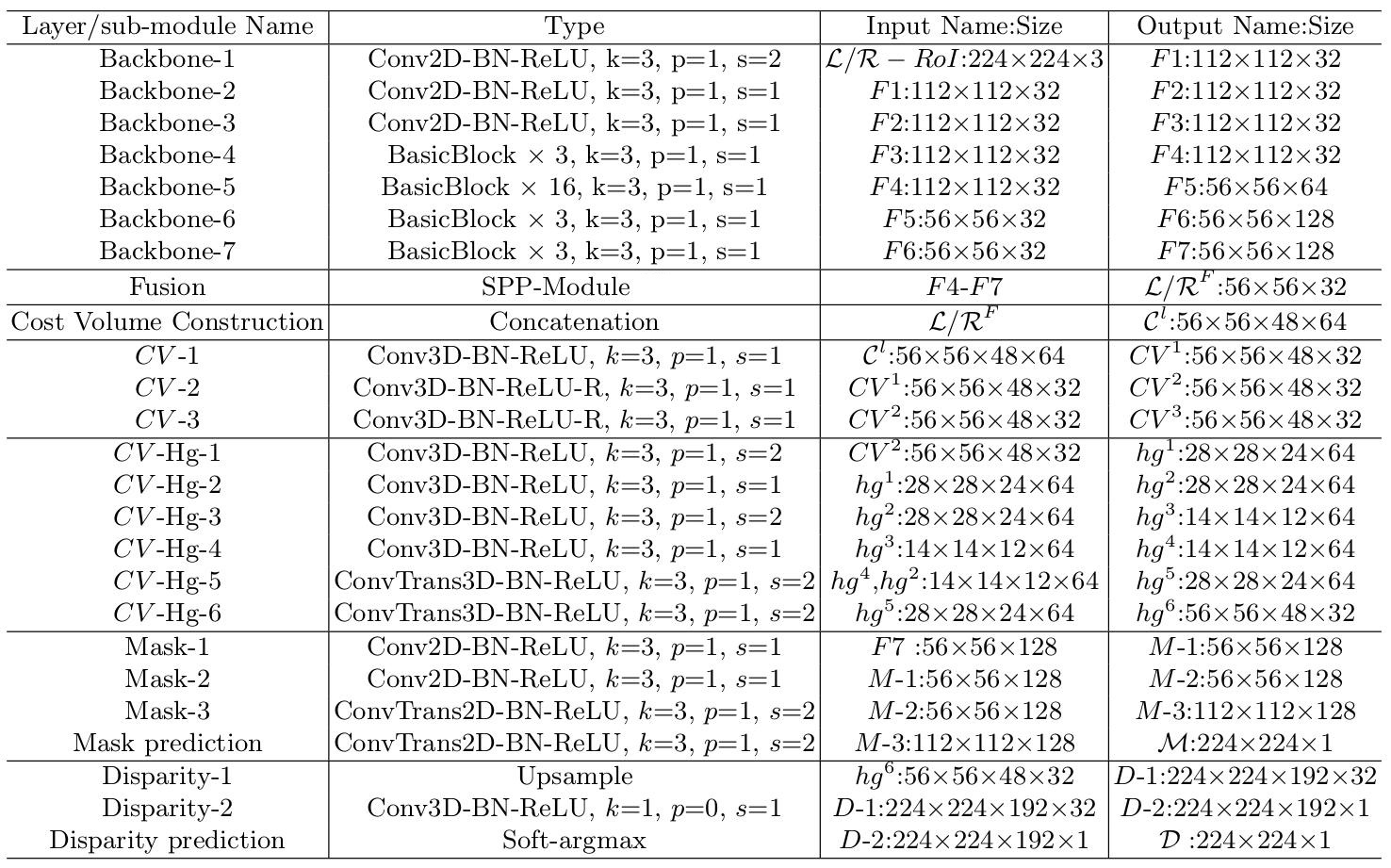}
			\caption{Detailed architecture of the visible surface extraction module $V$. Conv3D-BN-ReLU is a 3D convolution layer followed by batch normalization and ReLU activation. ConvTrans3D denotes a transposed 3D convolution layer. Conv3D-BN-ReLU-R means the output is added to the input as the residual. k denotes kernel size. p denotes padding. s denotes stride. d denotes dilation. SPP-Module refers to the pyramid module in~\cite{chang2018pyramid}.}
			\label{model_V}
		\end{center}
	\end{figure*}

\begin{figure*}[h!]
		\begin{center}
			\centering
			\includegraphics[width=0.7\linewidth, trim=0cm 0cm 0cm 0cm]{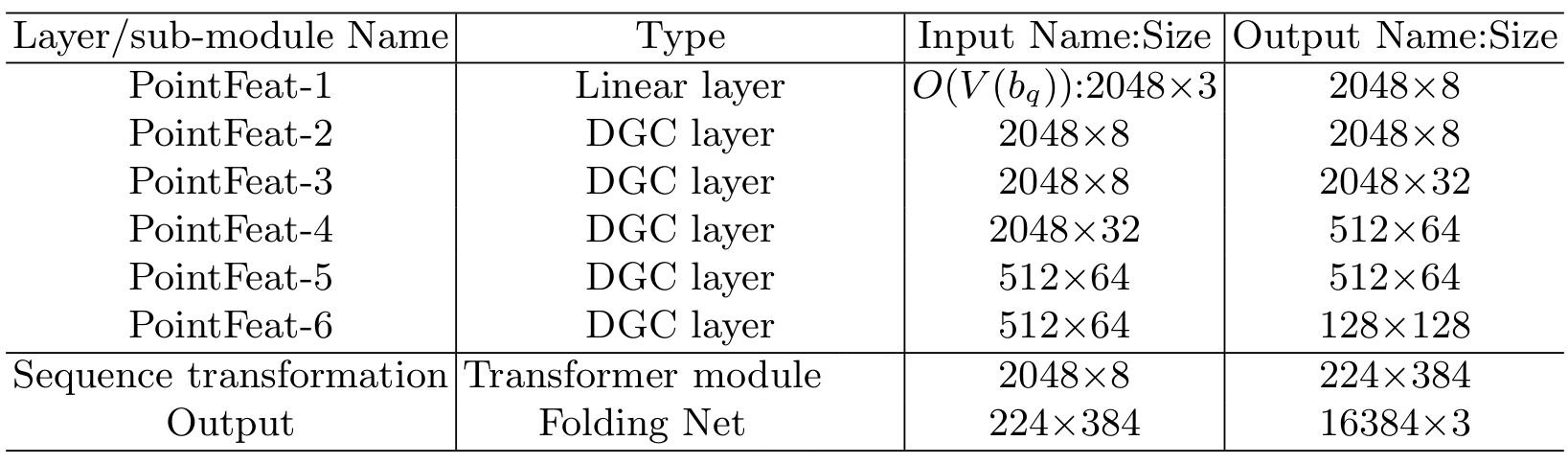}
			\caption{Detailed architecture of the unseen surface hallucination module $Ha$. The DGC layer denotes a dynamical graphical convolution layer. The transformer module and the folding net modules follows~\cite{yu2021pointr} and~\cite{yang2017foldingnet} respectively.}
			\label{model_Ha}
		\end{center}
	\end{figure*}

\begin{figure*}[h!]
		\begin{center}
			\centering
			\includegraphics[width=0.6\linewidth, trim=0cm 0cm 0cm 0cm]{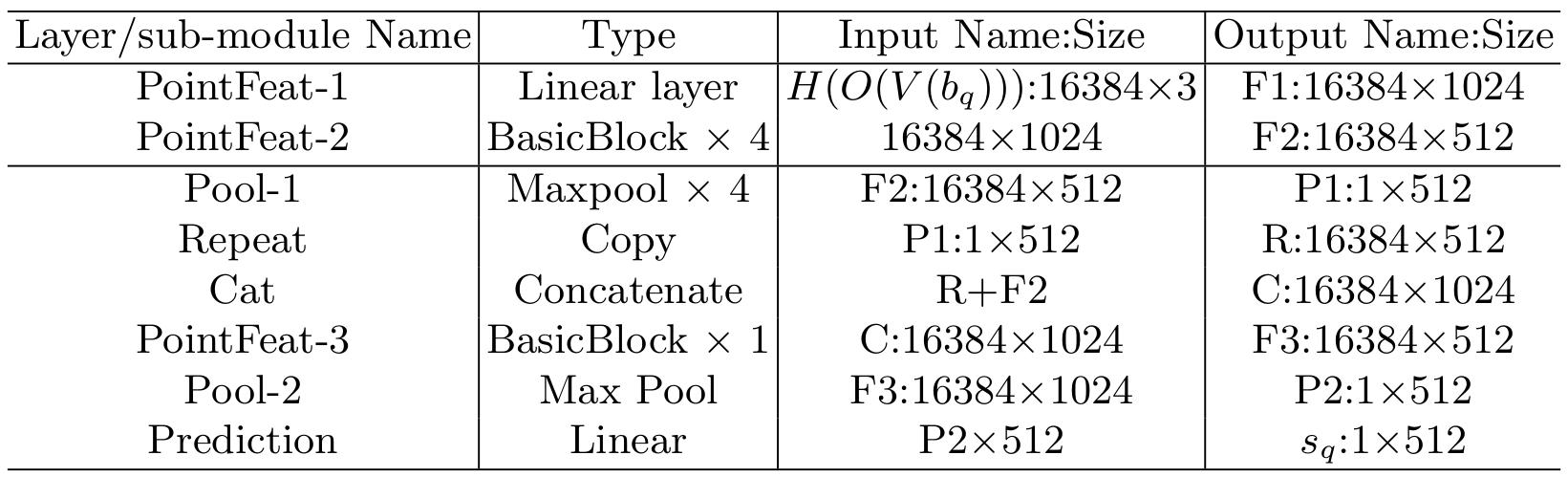}
			\caption{Detailed architecture of the point encoder module.}
			\label{model_En}
		\end{center}
	\end{figure*}

\begin{figure*}[h!]
		\begin{center}
			\centering
			\includegraphics[width=0.8\linewidth, trim=0cm 0cm 0cm 0cm]{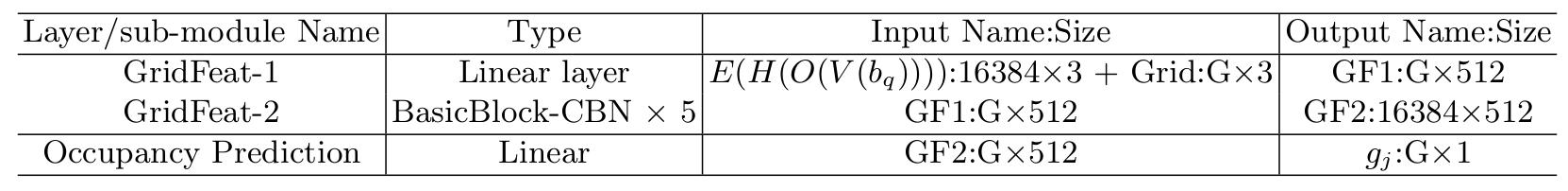}
			\caption{Detailed architecture of the shape decoder module. BasicBlock-CBN is a Resnet basic block with conditional batch normalization.}
			\label{model_De}
		\end{center}
	\end{figure*}

	The shape estimation branch in Ego-Net++ is visualized in Fig. 3 in the main text. It has learnable parameters $\{Ha, V, E\}$.  The detailed hyper-
	parameters of $V$ shown in Figure \ref{model_V}., which consists of 2D feature extraction, cost volume construction and processing, and depth/mask prediction. The detailed hyper-parameters of $Ha$, $E$, and the shape decoder are shown in Figure \ref{model_Ha}, Figure \ref{model_En}, and Figure \ref{model_De} respectively.
	
	\section{Training details}
	The supervision for Ego-Net++ consists of the orientation estimation part and the implicit shape estimation part. The training loss of the orientation estimation part is the same as Ego-Net~\cite{Li_2021_CVPR} and interested readers can refer to the official repository for more details. For the implicit shape estimation branch, the training loss consists of a cross-entropy segmentation loss $L_{seg}$, a smooth $L_{1}$ disparity estimation loss $L_{disp}$ following~\cite{sun2020disp}, and a hallucination loss $L_{Ha}$ as

	\begin{equation}
	L_{shape} = L_{seg} + L_{disp} + L_{Ha}.
	\end{equation}
	
	In implementation, we train $V$ and $Ha$ separately. $V$ is trained with $L_{ seg}$ + $L_{disp}$. We train with a batch size of 16 instances for 50 epochs. Adam optimizer is used and the learning rate is 0.001. For training $Ha$ we use ShapeNet training set as~\cite{yu2021pointr}. The hallucination loss is a Chamfer Distance loss between the predicted and ground truth point clouds. The training adopts a batch size of 50 and lasts 300 epochs. The learning rate starts at 0.001 and is multiplied by 0.9 after every 50 epochs. The experiments are conducted on NVIDIA RTX 3090 GPUs.
	
	The training process of the proposal model follows LIGA~\cite{guo2021liga}.
\bibliography{reference}
\end{document}